\documentclass[12pt]{article}

\usepackage[table]{xcolor}
\usepackage[margin=1in]{geometry}
\usepackage{amsmath,amssymb,amsthm,mathtools}
\usepackage{booktabs,array,enumitem,graphicx,microtype,subcaption}
\usepackage[section]{placeins}
\usepackage{float}
\usepackage[authoryear,round]{natbib}
\usepackage{algorithm,algorithmic}
\usepackage[colorlinks=true,allcolors=black]{hyperref}

\graphicspath{{figures/}}
\renewcommand{\epsilon}{\varepsilon}

\newcommand{\safeincludegraphics}[2][]{\IfFileExists{#2}{\includegraphics[#1]{#2}}{\fbox{\begin{minipage}[c][1.2in][c]{0.85\linewidth}\centering Missing figure file\end{minipage}}}}
\newcommand{\se}[1]{{\scriptsize$\pm$#1}}

\newtheorem{proposition}{Proposition}

\newtheorem{corollary}{Corollary}

\newcommand{\Y}{\mathcal Y}
\newcommand{\A}{\mathcal A}
\newcommand{\B}{\mathcal B}

\newcommand{\1}{\mathbf 1}
\newcommand{\E}{\mathbb E}
\newcommand{\R}{\mathbb R}
\newcommand{\Prb}{\mathbb P}
\newcommand{\clip}{\operatorname{clip}}
\newcommand{\Quantile}{\operatorname{Quantile}}
\newcommand{\argmin}{\operatorname*{argmin}}
\newcommand{\argmax}{\operatorname*{argmax}}
\newcommand{\TV}{\operatorname{TV}}

\title{Online Conformal Prediction Methods for Sequential Loss Minimization}
\author{
  Tao Wang,\,
  Tony Zhou,\, and
  Edgar Dobriban\thanks{University of Pennsylvania. E-mail addresses:
  \href{mailto:tawan@wharton.upenn.edu}{\nolinkurl{tawan@wharton.upenn.edu}},
  \href{mailto:zhou45@sas.upenn.edu}{\nolinkurl{zhou45@sas.upenn.edu}}, and
  \href{mailto:dobriban@wharton.upenn.edu}{\nolinkurl{dobriban@wharton.upenn.edu}}.
  TW and TZ contributed equally.}
}
\date{\today}

\begin{document}
\maketitle

\begin{abstract}
 In many
 online or sequential decision-making problems, 
 the goal is to minimize a 
 sum of loss functions by taking actions sequentially based on newly revealed features.
 We consider such a setting, where the loss depends on a class label that is revealed after the action, and where a probabilistic classifier is available.
 A natural approach is to consider a best response action that calculates the expected loss of each action and then chooses the optimal one.
 However, when the predictor is miscalibrated, this approach can lead to costly actions.
 Motivated by advances in conformal prediction, we instead consider a different approach, which is less sensitive to the accuracy of the probabilistic predictions.
 We
propose \emph{Decision-Aware Online Conformal Prediction} (DCP), a method to
 construct a sequence of prediction sets for the outcomes, and then choose actions robustly based on these sets.
 While there are many existing methods for online conformal prediction, these do not take the downstream actions and losses into consideration.
 We show that our methods satisfy long-term empirical coverage and certain bounded regret guarantees.
On four time series problems, where even powerful probabilistic classifiers can sometimes be inaccurate, DCP 
compares favorably in terms of decision loss to several baselines.
\end{abstract}

\tableofcontents

\section{Introduction}

As the prediction accuracy of machine learning models is constantly increasing, they are increasingly integrated in decision-making workflows.
 These decisions are often made downstream,
  leveraging the predictions of the machine learning algorithms. 
Often these decisions must be made sequentially, online, at each time step, as more and more data becomes available. 

We consider such a setting where we also have a machine learning model that predicts the probabilities of various outcomes at each time step. This model can be retrained periodically using the 
recently observed data. 
We also assume that there is a finite set of actions that we can choose from and given the unobserved outcome, each action and outcome  
lead to a well-defined loss (or equivalently, a specified reward).
For instance, in an online electricity market, a model may predict the probability of a price spike in an upcoming trading interval using recently observed prices and demand. Before the future price is observed, a consumer must choose whether to run as scheduled or curtail electricity consumption. In a simple cost model, curtailment incurs a fixed cost (for instance, the value of production forgone by temporarily idling equipment), regardless of whether a price spike occurs. Running as scheduled incurs a larger cost from purchasing electricity at higher prices if a spike occurs, and no additional cost beyond normal ones otherwise.

The natural starting point is to 
act according to the \emph{best response} (BR) action, 
choosing the action with the smallest expected loss under the predicted probabilities. 
If these probabilities are close to
the true conditional outcome probabilities, 
one would expect the BR action to nearly minimize the expected loss. 

In practice, however, a predictor may be inaccurate or miscalibrated, for example after a distribution shift, which can lead to incorrect and costly decisions. 
Even a somewhat accurate predictor can give poor decisions when its probabilities are near the threshold at which the best action changes.

An alternative approach uses prediction sets and has been studied in offline  settings, where a calibration sample is available before decisions are made on new observations \citep{kiyani2025decision,wang2026optimal}. A prediction set represents plausible outcomes, and a robust rule uses the set to choose an action
for a specified loss.

However, as we discussed, often decisions must be made in an online manner, in which case  the prediction sets 
constructed by leveraging 
the exchangeability from the offline setting may not have appropriate coverage guarantees.
 There is a variety of 
 methods for constructing prediction sets in an online manner, aiming for long-term coverage, even for arbitrary data streams, see e.g., 
\cite{gibbs2021adaptive,xu2023conformal,zaffran2022adaptive,bastani2022practical,gibbs2024conformal,angelopoulos2023conformal,bhatnagar2023improved,zhang2024discounted,podkopaev2024adaptive}, etc.
These methods construct prediction sets aiming to control the long-term average of the 
coverage errors $e_t=\1\{y_t\notin C_t\}$.
 They aim to ensure that the long-term fraction of errors is controlled at some level $\alpha$, i.e., 
 $T^{-1}\sum_t e_t\approx\alpha$.
One can then imagine taking actions in a worst-case robust manner according to these prediction sets. 

We 
therefore 
take a somewhat different perspective.
At each round, we consider a nested family of candidate sets indexed by a local level $\beta$.
A smaller local level is useful when enlarging the set improves the decision.
Of course, this choice must be learned online.
 Starting from this idea, we develop
a method called \emph{Decision-Aware Online Conformal Prediction} (DCP), which selects a local level $\beta_t$ by minimizing a surrogate for the loss of the action associated with the candidate set.
A scalar offset that shifts the thresholds of all candidate sets is updated from the observed errors to keep the long-run miscoverage at $\alpha$.

For selecting the action, \citet{wang2026optimal} minimize worst-case expected loss over distributions placing at most a specified mass (i.e., the miscoverage level) outside a prediction set. In the online setting, long-run coverage does not determine the probability mass outside a candidate set at a given round, so we need an additional controller that corrects the predicted out-of-set mass of each label using observed outcomes.

We provide two types of theoretical results for our method. 
First, we show that 
the time-averaged coverage of our method converges to the desired target at rate $O(1/T)$ on every data stream.
Second, we give an oracle inequality comparing the chosen action with any 
selection from the candidate actions formed using DCP's actual controller state. Against the Bayes-optimal action, an additional term measures the gap between the best current candidate and the best action in the full action space. 

 Since the choice of the surrogate is crucial,
we first consider DCP[$J$], which evaluates candidate
actions by their plug-in expected loss. This evaluation can inherit
errors from the original probabilities, motivating DCP[$V$], which
instead uses the corrected robust value to select the level.
For DCP[$J$] and DCP[$V$], we also bound the regret against the Bayes-optimal action in terms of the errors of the predicted probabilities and of the corrected out-of-set masses, respectively.
The robust value can be conservative, so we run both versions
and combine their actions using AdaHedge \citep{derooij2014follow}.
With immediate label feedback, the cumulative loss of this hedged method, averaged over its randomization, exceeds the smaller cumulative loss of the two runs by at most $O(\sqrt T)$; we show related bounds for the delayed feedback setting from the experiments in the appendix.

In our experiments, we compare our method with 
several natural baselines that use the predicted probabilities directly.
 In particular the best response (BR) method calculates the expected loss using the predicted probabilities and chooses an optimal action accordingly. 
Direct Loss Adaptation (DLA) corrects predicted action costs using past residuals and combines the corrected policies with BR by AdaHedge (Appendix~\ref{app:dla}).
 We also compare with the Unbiased Prediction (UP) method of \citet{noarov2025high}, which takes the best response to forecasts that are unbiased conditional on the action this best response selects.
 Finally, we compare with methods that first construct a prediction set and then take a robust action; for instance with the
  ACI \citep{gibbs2021adaptive} and DtACI \citep{gibbs2024conformal}
  prediction sets.
DCP has lower mean decision loss than the compared methods on all four datasets.

\subsection{Related Work}
\label{subsec:related-work}

{\bf Direct decision optimization and calibration.}
Our objective is to minimize the loss of sequential decisions. 
A natural starting
point is the best response (BR), which minimizes expected loss under the predicted
probabilities. Its performance depends on whether those probabilities accurately
represent the risks relevant to the decision. Decision calibration addresses this
issue by post-processing predictions so that, for every loss in a specified
class, the expected loss of the Bayes decision rule computed from the predictions
equals its actual expected loss \citep{zhao2021calibrating}. 
Decision calibration constrains forecast quality for downstream decisions, whereas our DLA baseline directly corrects forecasts of action losses using sequential residuals. 

Our work is also related to work on high-dimensional prediction for sequential decision making. 
In the online
adversarial setting, forecasts that are unbiased given
 a collection of
events, which may depend on the forecasts themselves, give diminishing swap
regret to downstream decision makers whose best responses define events in that
collection \citep{noarov2025high}. 
Our work shares the goal of minimizing cumulative downstream decision
loss, but studies this goal through prediction sets under a global
coverage constraint. Whereas their swap-regret guarantee compares
decisions with fixed remappings of the actions taken, we compare with
online rules that choose among the candidate actions
induced by our prediction sets, and with the Bayes-optimal action.

{\bf Online recalibration and related decision guarantees.}
 A sequence of forecasts can be recalibrated
 aiming to match predicted
probabilities with empirical ones. 
\citet{kuleshov2017estimating} develop a method based on online
calibration within bins of the original forecasts.
 In certain cases recalibrated probabilities can lead to a larger downstream loss under certain metrics. 
\citet{okoroafor2024faster} use Blackwell approachability to improve the
tradeoff between calibration error and excess loss under a proper scoring rule.
For binary outcomes and Lipschitz proper losses,
\citet{hu2026optimal} obtain sharper and sometimes optimal tradeoffs. 

The consequences of calibration for decision quality have also been studied
directly. For binary outcomes, \citet{hu2024calibration} introduce calibration
decision loss, which measures the largest improvement in decision payoff
obtainable by recalibrating forecasts over bounded decision problems, and
give an online algorithm with vanishing expected calibration decision loss.
Online omniprediction seeks a forecast sequence whose loss-specific best
responses compete with a prescribed predictor class across a family of losses
\citep{garg2024oracle,okoroafor2025near}.
\citet{okoroafor2025near} obtain near-optimal regret bounds for binary
outcomes and Lipschitz losses, while \citet{hu2026simultaneous} extend the
approach to multiclass outcomes.
However, 
these benchmarks differ from those in our analysis. Online omniprediction
compares with a specified predictor class, whereas 
our analysis of DCP
compares with sequences of candidate actions generated using
DCP's controller state, or with the conditional Bayes-optimal action.

{\bf Online conformal prediction.}
For background on conformal prediction, see
\citet{vovk1999machine,vovk2005algorithmic,angelopoulos2021gentle,angelopoulos2024theoretical}.
Under exchangeability, \citet{gibbs2025conditional} give finite-sample coverage guarantees
within each of a prespecified collection of subgroups. For regression with missing covariates,
\citet{zaffran2023missing} construct prediction intervals with coverage conditional on which covariates are missing.
For online data, adaptive conformal inference (ACI) updates a
miscoverage parameter so that realized miscoverage converges to a desired value \citep{gibbs2021adaptive}. Adaptive step-size aggregation
strengthens these guarantees under arbitrary distribution shifts
\citep{gibbs2024conformal}, while conformal PID control connects the updates to
proportional--integral--derivative control and models conformal scores
prospectively \citep{angelopoulos2023conformal}. 

For dependent time series, \citet{zaffran2022adaptive} aggregate ACI predictors
with different step sizes. In an application to French electricity spot prices,
\citet{dutot2024adaptive} combine conformal calibration with online aggregation to produce prediction intervals.
Other approaches minimize strongly adaptive regret, the worst-case regret over all intervals
of a fixed length, with approximately valid coverage \citep{bhatnagar2023improved},
or obtain threshold- and group-conditional empirical coverage against adversarial
sequences \citep{bastani2022practical}. \citet{areces2025online} derive the update
from an explicit online optimization problem, and \citet{liu2026universal} remove
its step size through a reduction to portfolio selection. 
Our focus is the use of online prediction sets for a specified downstream decision loss, with feedback correction of the labelwise masses entering the action rule. 

{\bf Conformal prediction for decisions and risk.}
Several papers tailor prediction sets or calibration to a downstream
objective. \citet{kiyani2025decision} study risk-averse agents who optimize a
quantile of the loss and show that, under a marginal coverage constraint, a
max--min policy over the set is optimal for that criterion.
\citet{wang2026optimal} instead minimize expected loss against the worst
distribution consistent with the coverage statement. 

\citet{cortesgomez2025utility} use the downstream loss to construct prediction sets
and apply their method to hierarchical diagnosis. \citet{lekeufack2024conformal}
calibrate the decision itself. Online adaptive risk control updates a scalar
threshold from past losses and gives a deterministic long-run risk guarantee
under arbitrary sequences \citep{feldman2023achieving}.
Localized adaptive risk
control replaces the scalar by a function in a reproducing kernel Hilbert space
\citep{zecchin2024localized}. In Bayesian optimization, \citet{kim2025robust} use
localized online conformal prediction to calibrate the Gaussian process surrogate that guides the choice of the next query.

\section{Problem setup}
\label{sec:setup}

\subsection{Online loss minimization and the best response}
\label{subsec:online-decisions}

Let $\Y$ be an outcome space and $\A$ be an action space, both finite and nonempty. The loss of action $a$ when the outcome is $y$ is $\ell(a,y)$, where $0\leq\ell(a,y)\leq\ell_{\max}=\max_{y,a}\ell(a,y)$ and $\ell_{\max}>0$. At round $t$, the learner observes covariates $x_t$.
The learner has access to a predictor that can output
predicted probabilities $\widehat p_t$ 
of the classes in $\Y$, using the entire history of observations.
A practical way to implement this is to maintain a probabilistic classifier that may be updated using past observations.

Using all the available data, the learner
chooses an action $a_t\in\A$.
It then observes the label $y_t$ and incurs loss $\ell(a_t,y_t)$.  Our objective is online loss minimization: choosing actions to keep the cumulative loss $\sum_{t=1}^T\ell(a_t,y_t)$ small.

A direct approach to using the predictor is the plug-in best response (BR) action,
\begin{equation}
a_t^{\rm BR}\in\argmin_{a\in\A}\sum_{y\in\Y}\widehat p_t(y)\ell(a,y),
\label{eq:best-response}
\end{equation}
with a fixed rule for ties. If $\widehat p_t$ 
approximates the distribution of $Y_t$ given the observed data, this action should approximately minimize the conditional expected loss. 
Otherwise, best response can lead to incorrect decisions,
as we will see 
in our empirical examples. 

To give some intuition, consider a simple toy one-round example where 
$\Y=\A=\{0,1\}$, where $y=1$ denotes an external hazard (e.g., a storm) and $a=1$ denotes ``protection'' (e.g., deploying a protective barrier). Suppose an adverse event causes damage equal to unity, while protection prevents this damage at cost $f\in(0,1)$. Subtracting the best achievable cost for each outcome and dividing by $f$ gives the loss matrix
$
\bigl(\ell(a,y)\bigr)_{a\in\A,\,y\in\Y}=
\begin{pmatrix}
0&(1-f)/f\\
1&0
\end{pmatrix}.
$
Thus, unnecessary protection incurs loss equal to unity, while a missed event incurs loss $(1-f)/f$.
The zero diagonal represents no excess cost relative to the best action in hindsight. 

Writing $\widehat p_t=\widehat p_t(1)$, 
with ties broken toward no protection, BR chooses the protective action 
when $\widehat p_t>f$. 
If the cost is $f=0.10$, a forecast of $0.15$ therefore triggers protection. If the true conditional event probability is $0.05$, however, the expected loss is $0.95$ with protection versus $0.45$ without it. 
Therefore, a  predictor that provides incorrect probabilities can 
induce costly decisions with BR, even if those probabilities are not too far from correct.
BR is especially sensitive to predictions that cross certain decision boundaries (equal to $f$ above).  

\subsection{An alternative: actions from prediction sets}
\label{subsec:sets-actions}

Motivated by the above sensitivity of BR to certain precise numerical predictions, 
we investigate an alternative approach. 
Instead of basing our decisions directly on the exact numerical predictions, we consider an intermediate structure, namely constructing sets of predicted outcomes that are plausible.
Such prediction sets have been investigated in statistics since the 1940s, see e.g., \cite{Wilks1941,Wald1943}, and have more recently been studied extensively in the area of conformal prediction, including in online settings, see e.g., \cite{vovk2005algorithmic,gibbs2021adaptive,xu2023conformal,angelopoulos2023conformal}, and the works cited in our related work. 

Specifically, we construct a prediction set, 
and choose an action informed by it.
 As is standard in conformal prediction, 
the learner forms scores $s_t(y)\in[0,1]$, 
which  
measure the degree to which the outcome $y$ is viewed as unusual. 
A popular score is $s_t(y)=1-\widehat p_t(y)$, so labels with smaller predicted probability are viewed as more unusual. 
The learner also forms 
a prediction set $C_t\subseteq\Y$. 
The size of the sets can be tuned online aiming to capture a certain fraction of the outcomes.
 Specifically, 
after observing the outcome $y_t$, 
the learner can evaluate
the miscoverage indicator $e_t=\1\{y_t\notin C_t\}$. 
As described below, the size of the sets can be tuned  
so that the time-averaged miscoverage approaches a prescribed $\alpha\in(0,1)$. 

To specify how to use the prediction sets to choose actions,
 we take inspiration from the offline setting studied by \citet{wang2026optimal}, who developed a robust loss and decision-making rule based on prediction sets. 
We first review their approach here, and then explain how we adapt it. 
For a prediction set $C\subseteq\Y$ and $a\in\A$, define 
the \emph{within-set worst-case loss} 
$\ell_{\rm in}(C,a)=\max_{y\in C}\ell(a,y)$ and 
the \emph{out-of-set worst-case loss} 
$\ell_{\rm out}(C,a)=\max_{y\notin C}\ell(a,y)$, with the conventions $\ell_{\rm in}(\emptyset,a)=0$ and $\ell_{\rm out}(\Y,a)=0$. 
For a scalar 
$\beta\in[0,1]$, 
the robust loss of \citet{wang2026optimal} is
\begin{equation}
L(C,a;\beta)=\ell_{\rm in}(C,a)+\beta\big[\ell_{\rm out}(C,a)-\ell_{\rm in}(C,a)\big]_+.
\label{eq:robust-loss}
\end{equation}
 To interpret this,  
 for a non-empty set
 $C$, $L(C,a;\beta)$ equals the supremum of 
the expected loss 
 $\E_{Y\sim Q}[\ell(a,Y)]$ over distributions $Q$ 
 that are consistent with the marginal coverage constraint
 $Q(C)\geq1-\beta$. 
 The first term  
 in \eqref{eq:robust-loss}
 accounts for the worst-case outcome being inside the set; the second accounts for a possible increase in loss when the worst outcome is outside the set.
 At $\beta=0$, the rule reduces to the minimax action over $C$ \citep{kiyani2025decision}. 
 Once we have this robust loss, we can choose an action by simply minimizing it, i.e., 
\begin{equation}
a(C,\beta)\in\argmin_{a\in\A}L(C,a;\beta).
\label{eq:robust action}
\end{equation}

\subsection{Online conformal prediction}
\label{subsec:aci}

Since our goal is minimizing the loss in an online setting, 
and we aim to approach this by first constructing prediction sets online,
 we now review relevant background material about online conformal prediction. 
We will then discuss how to adapt these methods to our setting of loss minimization.

 A basic method for online 
 prediction set construction called 
adaptive conformal inference (ACI; \citealp{gibbs2021adaptive})  maintains a level $\alpha_t$ 
at each time step, 
and updates it using sub-gradient descent on the quantile regression (aka pinball) loss.
With step size $\gamma>0$, the updates starting from some $\alpha_1 \in [0,1]$ take the form
\begin{equation}
\alpha_{t+1}=\alpha_t+\gamma(\alpha-e_t).
\label{eq:aci-update}
\end{equation}
This update uses the miscoverage indicator computed from the observed label.
When the reported indicator can be flipped, \citet{wang2026corrupted} bound the effect
of these errors on long-run miscoverage and propose corrections to the update.

For $0<\alpha_t<1$, the set is $C_t=\{y\in\Y:s_t(y)\leq\Quantile_{1-\alpha_t}(H_{t-1})\}$. Here $H_0$ is a nonempty finite multiset of initial scores in $[0,1]$, and $H_{t-1}$ is obtained by appending the observed scores, retaining their multiplicities. For a multiset of size $n$ and $u\in(0,1)$, $\Quantile_u$ denotes its $\lceil un\rceil$-th order statistic. We take $C_t=\Y$ when $\alpha_t\leq0$ and $C_t=\emptyset$ when $\alpha_t\geq1$. 

In our setting we will allow the learner to choose a local level $\beta_t$ from a nonempty finite grid $\B\subset(0,1)$ before observing $y_t$, form $C_t(\beta_t)$, and take its associated action $a_t(\beta_t)$. 
We will aim to choose the local levels $\beta_t$ to approximately minimize the online loss while maintaining coverage.
Since the losses are unknown when the levels are chosen,
we will use loss surrogates to compare levels.

\section{Decision-aware online conformal prediction }
\label{sec:method}

Our method
DCP has three main steps in order to select the local level $\beta_t$ and its associated action. 
It first constructs candidate sets by shifting empirical quantiles by an offset, then assigns an action to each set using 
estimated out-of-set masses, and finally chooses a level by minimizing a decision surrogate. 
We explain each component in turn. 

For each candidate local level $\beta$ in the grid $\B$, 
as in ACI \citep{gibbs2021adaptive},
let $q_t^0(\beta)=\Quantile_{1-\beta}(H_{t-1})$ denote the empirical quantile of the score history.
We 
will  use a scalar offset $b_t$ to form $C_t(\beta;b_t)=\{y\in\Y:s_t(y)\leq q_t^0(\beta)+b_t\}$.
 The scalar offset is learned online
aiming to keep the long-run miscoverage at $\alpha$.

\subsection{Out-of-set mass and chosen action}
\label{subsec:surrogate}

To choose the local level $\beta_t$, we first aim to estimate 
the out-of-set probability mass 
for $C_{t,\beta}=C_t(\beta;b_t)$. 
A naive estimate would be 
$m^{\rm naive}_{t,y}(\beta)=\1\{y\notin C_{t,\beta}\}\widehat p_t(y)$.
However, since the predicted probabilities $\widehat p_t$ may be far from the truth, we consider correcting 
these estimates with 
a correction $d_{t,y}(\beta)\in[-c_{\max},c_{\max}]$.
Therefore we construct the 
estimate $\bar m^{\rm out}_{t,y}(\beta)=\1\{y\notin C_{t,\beta}\}\widehat p_t(y)+d_{t,y}(\beta)$. 
In order to ensure that these estimates are compatible with the candidate sets,
we project them by a map
$\Pi_t$, specified in Appendix~\ref{app:implementation}, 
enforcing $\widehat m^{\rm out}_{t,y}(\beta)=0$ for $y\in C_{t,\beta}$,
as well as that they have nonnegative entries whose sum is at most one, and that they are nondecreasing in $\beta$ for each label. 
We thus obtain 
\begin{equation}
\{\widehat m^{\rm out}_{t}(\beta):\beta\in\B\}
=
\Pi_t\{\bar m^{\rm out}_{t}(\beta):\beta\in\B\}.
\label{eq:projection}
\end{equation}

The DCP action then 
minimizes the worst-case expected loss over distributions $Q$ satisfying $Q(y)\leq\widehat m^{\rm out}_{t,y}(\beta)$ for every $y\notin C_{t,\beta}$,
using 
more refined
per-label constraints instead of the single constraint $Q(C_{t,\beta})\geq1-\beta$ underlying \eqref{eq:robust-loss}.
This amounts to\footnote{Strictly speaking, the above interpretation of the loss function holds only 
for nonempty $C_{t,\beta}$; and the objective remains a heuristic when this set is empty.}  (Proposition~\ref{prop:calibrated-action-minimax} in Appendix~\ref{app:calibrated-action})
\begin{equation}
a_t(\beta)\in\argmin_{a\in\A}
\left\{
\ell_{\rm in}(C_{t,\beta},a)
+
\sum_{y\notin C_{t,\beta}}
\widehat m^{\rm out}_{t,y}(\beta)
\left[\ell(a,y)-\ell_{\rm in}(C_{t,\beta},a)\right]_+
\right\},
\label{eq:corrected-action}
\end{equation}
with ties broken by a fixed order on $\A$. 

For $C_{t,\beta}=\emptyset$, we use the same algebraic rule with $\ell_{\rm in}(\emptyset,a)=0$. 

The corrections are learned from the observed labels. After $y_t$ is observed, we record for every  pair of 
levels and labels 
whether $y_t$ equals the label 
and whether $y_t$ is outside of the set:
\begin{equation}
Z_{t,y}(\beta)=\1\{y_t=y,\ y_t\notin C_t(\beta;b_t)\},
\qquad (\beta,y)\in\B\times\Y.
\label{eq:Z}
\end{equation}
We then
update 
the correction
$d_{t+1,y}(\beta)$
so that 
underestimated outside occurrences increase it,
and vice versa.
Specifically,
for a learning rate $\eta_m$, we update
$d_{t+1,y}(\beta)=\clip_{[-c_{\max},c_{\max}]}[d_{t,y}(\beta)+\eta_m\{Z_{t,y}(\beta)-\widehat m^{\rm out}_{t,y}(\beta)\}]$. 
This feedback is designed to reduce persistent errors in the masses used by the action rule. 

\subsection{Algorithm}
\label{subsec:algorithm}

Once the candidate actions have been constructed, the remaining decision is which level to select. 
The expected losses 
associated with the actions and respective levels
are unknown, so DCP evaluates a surrogate $J_t(\beta)$ for the cost of $a_t(\beta)$. 
We consider two surrogates.

 The first
    is the plug-in objective, the expected loss of the DCP action under the predicted probabilities,
    \begin{equation}
    J_t(\beta)
    =
    \frac{1}{\ell_{\max}}
    \sum_{y\in\Y}\widehat p_t(y)\ell(a_t(\beta),y),
    \label{eq:J}
    \end{equation}
    leading to DCP[$J$].

We also consider a second objective, the robust value, given by the normalized optimal value of the program defining $a_t(\beta)$ in \eqref{eq:corrected-action},
    \begin{equation}
    V_t(\beta)
    =
    \frac{1}{\ell_{\max}}
    \min_{a\in\A}
    \left\{
    \ell_{\rm in}(C_{t,\beta},a)
    +
    \sum_{y\notin C_{t,\beta}}
    \widehat m^{\rm out}_{t,y}(\beta)
    \left[\ell(a,y)-\ell_{\rm in}(C_{t,\beta},a)\right]_+
    \right\},
    \label{eq:VLW}
    \end{equation}
    leading to DCP[$V$].

    The surrogates differ in how they use the probability estimates.
The plug-in objective $J_t$ evaluates
candidate actions under $\widehat p_t$, 
so it can  inherit
certain errors from these probabilities.
The mass corrections affect the candidate actions
but do not correct the probabilities used to evaluate them.
Thus, DCP[$J$] can select a poor local level even when the
out-of-set masses used by its action rule are accurate.

The robust value $V_t$ uses $\widehat p_t$ only through the sets
and corrected masses.
Using this value for level selection 
ensures DCP[$V$] uses the same loss
surrogate for action selection and level selection.
This value can nevertheless be conservative: it can overestimate the
expected loss and change the ranking of candidate levels.

The two criteria can therefore favor different
levels, and their relative decision loss can vary across
data streams.
Algorithm~\ref{alg:dcp} gives the complete procedure.
Later, we will also
 describe a practical choice which entails 
 using a hedging algorithm
(AdaHedge; \citealp{derooij2014follow}) to select between their actions, with weights determined
by past decision losses.

\begin{algorithm}[H]
\caption{\footnotesize Decision-Aware Online Conformal Prediction (DCP)}
\label{alg:dcp}\footnotesize
\begin{algorithmic}[1]
\REQUIRE target $\alpha\in(0,1)$, nonempty finite grid $\B\subset(0,1)$, step sizes $\eta_b,\eta_m>0$, correction bound $c_{\max}>0$, nonempty initial score multiset $H_0\subset[0,1]$, loss function $\ell$, and a pre-outcome surrogate $J_t$.
\STATE Initialize $b_1=0$ and $d_{1,y}(\beta)=0$ for all $(\beta,y)\in\B\times\Y$.
\FOR{$t=1,2,\ldots,T$}
\STATE Observe $x_t$ and compute $\widehat p_t$ and $s_t(y)$ for all $y\in\Y$.
\STATE For every $\beta\in\B$, form $C_t(\beta;b_t)=\{y:s_t(y)\leq q_t^0(\beta)+b_t\}$.
\STATE Form $\bar m^{\rm out}_{t,y}(\beta)=\1\{y\notin C_t(\beta;b_t)\}\widehat p_t(y)+d_{t,y}(\beta)$ for all $(\beta,y)$ and project to $\widehat m^{\rm out}_{t}(\beta)$ using $\Pi_t$.
\STATE For every $\beta\in\B$, compute $a_t(\beta)$ and evaluate $J_t(\beta)$.
\STATE Select $\beta_t\in\argmin_{\beta\in\B}J_t(\beta)$; among tied levels, take the largest if $b_t<0$ and the smallest otherwise.
\STATE Deploy $C_t=C_t(\beta_t;b_t)$ and $a_t=a_t(\beta_t)$.
\STATE Observe $y_t$, set $e_t=\1\{y_t\notin C_t\}$, and form $H_t$ by appending $s_t(y_t)$ to $H_{t-1}$.
\STATE Update each correction by $d_{t+1,y}(\beta)=\clip_{[-c_{\max},c_{\max}]}\big[d_{t,y}(\beta)+\eta_m\{Z_{t,y}(\beta)-\widehat m^{\rm out}_{t,y}(\beta)\}\big]$, with $Z_{t,y}(\beta)$ as in \eqref{eq:Z}.
\STATE Update $b_{t+1}=b_t+\eta_b(e_t-\alpha)$.
\ENDFOR
\end{algorithmic}
\end{algorithm}

Asymptotic coverage holds regardless of the choice of surrogate.
However, to guarantee that
the decision loss is small, we need the surrogate to be accurate, 
in a sense made explicit in Section~\ref{subsec:decision loss-regret}. 

Ordinary coverage requires the realized miscoverage frequency to approach
$\alpha$. We can enforce this condition directly through the offset $b_t$ while choosing the local
level to minimize the decision surrogate: $b_t$ increases after a miss, which enlarges all candidate sets, and decreases otherwise.
Several levels can have the same surrogate value; under the plug-in objective, for instance, this occurs whenever they lead to the same candidate action. DCP breaks ties using past coverage errors. Since $b_1=0$, we have $b_t=\eta_b\sum_{s<t}(e_s-\alpha)$, so $b_t<0$ means that the number of misses so far is below $\alpha(t-1)$. In this case DCP takes the largest tied level, whose set is the smallest; otherwise it takes the smallest tied level. With delayed labels, the rule also accounts for the rounds whose labels have not yet been observed (Appendix~\ref{app:delays}).

{\bf Discussion of the computational cost.}
Apart from predictor evaluation and score history maintenance, the decision step costs $O(|\B||\A||\Y|)$ and the projection costs $O(|\B||\Y|)$. 
One may also need to 
refit
the predictor. 
Storing the full score history requires memory linear in the number of observed scores, in addition to $O(|\B||\Y|)$ correction states. The order statistic used for the score quantiles and the computation of $\Pi_t$ are given in Appendix~\ref{app:implementation}, and the remaining implementation choices in Appendix~\ref{app:protocol}.

\section{Theoretical analysis}
\label{sec:theory}

In this section, we provide theoretical results on the performance of our methods. 
Proofs are in Appendix~\ref{app:proofs}.
Unless stated otherwise, the results in this section concern one prediction per round with its label available before the next round. Proposition~\ref{prop:delayed-coverage} treats delayed and batched coverage separately. 

As a starting point,
 we remark that 
our method achieves 
asymptotic coverage regardless of the accuracy of the predictors involved. 
In fact since the updates on $b_t$ are identical to those for ACI \citep{gibbs2021adaptive}, it follows directly from the results of \cite{gibbs2021adaptive} that 
\begin{equation}
\left|\frac1T\sum_{t=1}^T e_t-\alpha\right|
\le\frac{1/\eta_b+\max\{\alpha,1-\alpha\}}{T}.
\label{eq:greedy-direct-coverage}
\end{equation}
In particular, the DCP prediction sets satisfy asymptotic $1-\alpha$ coverage.

\subsection{Decision loss regret}
\label{subsec:decision loss-regret}

Of course, the main goal is to achieve a small total loss in an online fashion.
 We will now argue 
 that our algorithm can do this under some reasonable conditions. 
Let $(\Omega,\mathcal H,\mathbb P)$ be a probability space 
that includes 
the covariate--label process $(X_t,Y_t)_{t\geq1}$, any initial training data, and any algorithmic randomization.  The coverage bound \eqref{eq:greedy-direct-coverage} holds conditional on every realized sequence.

Let $\mathcal F_{t-1}\subseteq\mathcal H$ be the sigma-algebra generated by the information available before observing $X_t$.
Let $\mathcal G_t\subseteq\mathcal H$ be the sigma-algebra generated by $\mathcal F_{t-1}$, $X_t$, and any additional randomization, before $Y_t$ is revealed. 
Thus $b_t$ and $d_t$ are $\mathcal F_{t-1}$-measurable, and $C_t(\beta;b_t)$, $\widehat m^{\rm out}_t(\beta)$, $a_t(\beta)$, $J_t(\beta)$, and $\beta_t$ are $\mathcal G_t$-measurable. 
Write $a_{t,\beta}=a_t(\beta)$. 

Recalling $0\leq\ell(a,y)\leq\ell_{\max}$, write $\bar\ell(a,y)=\ell(a,y)/\ell_{\max}$ and define the normalized conditional risks 
\begin{equation}\label{risks}
r_t(a)=\mathbb E[\bar\ell(a,Y_t)\mid\mathcal G_t]\textnormal{ and } R_t(\beta)=r_t(a_{t,\beta}).
\end{equation}
For a fixed action $a$, the expectation in \eqref{risks} averages over the conditional distribution of the unobserved label $Y_t$ under $\mathbb P$, given $\mathcal G_t$. 

 In our analysis we will show that DCP achieves small regret relative to the best candidate action under a sufficient condition that requires the surrogate to be 
 a good approximation of the decision risk.
 Clearly since the surrogates play a crucial role in determining which actions to choose, it is reasonable that the regret depends on their accuracy compared to the true conditional averages $R_t$ that an oracle would use. 
To measure the error relevant to comparing
candidates, let
\begin{equation}
\omega_t=
\max_{\beta\in\mathcal B}\{R_t(\beta)-J_t(\beta)\}
-\min_{\beta\in\mathcal B}\{R_t(\beta)-J_t(\beta)\}.
\label{eq:greedy-error-span}
\end{equation}
A fixed additive error in all surrogate values does not affect
$\omega_t$. 
Under the uniform validity condition
$\max_{\beta\in\mathcal B}|J_t(\beta)-R_t(\beta)|\le\epsilon_t$, we have
$\omega_t\le2\epsilon_t$.
For the plug-in surrogate of DCP[$J$], one may take $\epsilon_t=\TV(\widehat p_t,p_t^*)=\frac12\|\widehat p_t-p_t^*\|_1$, where $p_t^*$ is the conditional label distribution given $\mathcal G_t$ (Proposition~\ref{prop:surrogate-validity} in Appendix~\ref{app:proof-surrogate-validity}).

 We now present a simple regret bound for DCP, 
with a proof in Appendix~\ref{app:proof-greedy-direct}. 

\begin{proposition}[Expected decision loss regret of DCP]
\label{thm:greedy-direct}
For every sequence of $\mathcal G_t$-measurable levels
$\widetilde\beta_t\in\mathcal B$, DCP, with ties in the selection of $\beta_t$ resolved by any $\mathcal G_t$-measurable rule, satisfies
\begin{equation}
\mathbb E\sum_{t=1}^T
\{\ell(a_{t,\beta_t},Y_t)-\ell(a_{t,\widetilde\beta_t},Y_t)\}
\le\ell_{\max}\mathbb E\sum_{t=1}^T\omega_t.
\label{eq:greedy-candidate-regret}
\end{equation}
\end{proposition}

Under uniform validity, the bound is at most $2\ell_{\max}\mathbb E\sum_{t=1}^T\epsilon_t$.
This bound
yields sublinear expected regret against every sequence of $\mathcal G_t$-measurable levels if $\mathbb E\sum_{t=1}^T\omega_t=o(T)$. 
The proposition applies to DCP[$J$] and DCP[$V$] with their respective surrogate errors.

We also combine DCP[$J$] and DCP[$V$] using AdaHedge \citep{derooij2014follow}, as in Section~\ref{sec:experiments}. Let $a_t^J$ and $a_t^V$ be the actions proposed by the two runs. AdaHedge assigns them nonnegative selection weights $w_t^J$ and $w_t^V$, computed from feedback available before prediction, and selects $I_t\in\{J,V\}$ according to
\begin{equation}
\mathbb P(I_t=k\mid\mathcal G_t)=w_t^k,
\qquad w_t^J+w_t^V=1,
\qquad k\in\{J,V\}.
\label{eq:dcp-mixture-weights}
\end{equation}
Here $\mathcal G_t$ includes the information of both runs before the fresh draw $I_t$, and the deployed action is $a_t^{I_t}$. The explicit weight formula under immediate feedback is given in \eqref{eq:adahedge-weights} in Appendix~\ref{app:combination}; the delayed-feedback update protocol is specified in Appendix~\ref{app:delays}.
With immediate feedback, the cumulative loss averaged over the draws of AdaHedge exceeds the smaller cumulative loss of the two runs by at most $O(\sqrt T)$ on every data stream (Proposition~\ref{prop:combination} in Appendix~\ref{app:combination}).

\subsection{Regret against the Bayes-optimal action}
\label{sec:greedy-external}

Proposition~\ref{thm:greedy-direct} compares DCP with any online rule that
chooses among the same candidate actions. A stronger benchmark is the Bayes-optimal
action, which minimizes the conditional expected loss over the entire
action space. 
Recalling the 
normalized conditional risks 
$R_t$ and $r_t$ from \eqref{risks},
let $r_t^{\rm cand}=\min_{\beta\in\mathcal B}R_t(\beta)$ be the
minimum conditional risk among the candidate actions, and
$r_t^\star=\min_{a\in\mathcal A}r_t(a)$ be the minimum conditional risk
over the entire action space. Choose a
$\mathcal G_t$-measurable Bayes-optimal action
$a_t^\star\in\operatorname*{argmin}_{a\in\mathcal A}r_t(a)$.
Define $\Gamma_t=r_t^{\rm cand}-r_t^\star\ge0$, the approximation error
from restricting decisions to the candidate actions.
Proposition~\ref{thm:greedy-direct} implies the following corollary.

\begin{corollary}
\label{cor:greedy-bayes}
Under the conditions of Proposition~\ref{thm:greedy-direct}, the regret of DCP
against the Bayes-optimal action satisfies
\begin{equation}
\mathbb E\sum_{t=1}^T
\{\ell(a_{t,\beta_t},Y_t)-\ell(a_t^\star,Y_t)\}
\le\ell_{\max}\mathbb E\sum_{t=1}^T(\omega_t+\Gamma_t).
\label{eq:greedy-bayes-gap-bound}
\end{equation}
\end{corollary}

Here $\omega_t$ measures the surrogate error in comparing the candidate
actions, and $\Gamma_t$ measures how much worse the best of them is than
the Bayes-optimal action.
In particular, $\mathbb E\sum_{t=1}^T(\omega_t+\Gamma_t)=o(T)$ is sufficient for sublinear expected regret against the Bayes-optimal action. 

For DCP[$J$], the regret can also be bounded through the predicted
probabilities and their error. For every action, define its plug-in risk
$\widehat r_t(a)=\sum_y\widehat p_t(y)\bar\ell(a,y)$. DCP[$J$]
uses
the plug-in surrogate
$J_t(\beta)=\widehat r_t(a_{t,\beta})$. The quantity
$\widehat\Gamma_t=\min_{\beta\in\mathcal B}\widehat r_t(a_{t,\beta})
-\min_{a\in\mathcal A}\widehat r_t(a)\ge0$
is the plug-in counterpart of $\Gamma_t$ and is available before
observing $Y_t$. It vanishes whenever the candidates
contain a plug-in best response.
 Then we have the following regret bound for DCP[$J$],
 which depends
on the expected error $\mathbb E\|p_t^*-\widehat{p}_t\|_1$ of the predicted probabilities.
The proof is in Appendix~\ref{app:greedy-external}.

\begin{proposition}
\label{prop:greedy-plugin-bayes}
DCP[$J$] satisfies
\begin{equation}
\mathbb E\sum_{t=1}^T
\{\ell(a_{t,\beta_t},Y_t)-\ell(a_t^\star,Y_t)\}
\le\ell_{\max}\mathbb E\sum_{t=1}^T
(\widehat\Gamma_t+\|p_t^*-\widehat{p}_t\|_1).
\label{eq:greedy-plugin-bayes}
\end{equation}
\end{proposition}

 DCP[$V$] instead evaluates
candidates by the robust value, which uses $\widehat p_t$ only through
the candidate sets and the corrected out-of-set masses. For a candidate
level $\beta$ and an action $a$, even with exact out-of-set masses, the
objective in \eqref{eq:corrected-action}, divided by $\ell_{\max}$, can
exceed $r_t(a)$, with the excess
\begin{equation}
\kappa_t(\beta,a)=\frac1{\ell_{\max}}\sum_{y\in\Y}p_t^*(y)
\bigl[\ell_{\rm in}(C_{t,\beta},a)-\ell(a,y)\bigr]_+.
\label{eq:greedy-conservatism}
\end{equation}
The errors of the out-of-set masses are measured by
\begin{equation}
\rho_t^-(\beta)=\sum_{y\notin C_{t,\beta}}
[p_t^*(y)-\widehat m^{\rm out}_{t,y}(\beta)]_+,
\qquad
\rho_t^+(\beta)=\sum_{y\notin C_{t,\beta}}
[\widehat m^{\rm out}_{t,y}(\beta)-p_t^*(y)]_+.
\label{eq:greedy-mass-errors}
\end{equation}
 Then we have the following regret bound for DCP[$V$],
 which depends
 on the above error measures. 
 The proof is in Appendix~\ref{app:greedy-external}. 
\begin{proposition}
\label{prop:greedy-robust-bayes}
For DCP with any surrogate,
$\Gamma_t\le\min_{\beta\in\mathcal B}\{\kappa_t(\beta,a_t^\star)+\rho_t^-(\beta)+\rho_t^+(\beta)\}$
at every round. Moreover, DCP[$V$] satisfies
\begin{equation}
\mathbb E\sum_{t=1}^T
\{\ell(a_{t,\beta_t},Y_t)-\ell(a_t^\star,Y_t)\}
\le\ell_{\max}\mathbb E\sum_{t=1}^T
\Bigl[\rho_t^-(\beta_t)+\min_{\beta\in\mathcal B}
\{\kappa_t(\beta,a_t^\star)+\rho_t^+(\beta)\}\Bigr].
\label{eq:greedy-V-bayes}
\end{equation}
\end{proposition}

The approximation terms in the above results,
such as $\Gamma_t$, reflect the choice of
procedure and cannot in general be eliminated by a sharper analysis.
Even when the predicted probabilities are exact, DCP can have linear
regret against a comparator that uses the same robust action rule and
satisfies the same expected coverage requirement
(Appendix~\ref{app:approximation-necessity}).
 However, these oracle comparisons do not actually lead to a practically useful alternative for DCP, and mostly have value in terms of revealing insights and conditions under which the algorithm is expected to perform well. 
 In practice, when we compare our algorithm with a variety of baselines in the next section, we will observe its empirical advantages in more detail. 

\section{Experiments}
\label{sec:experiments}

We now present experiments to evaluate the empirical performance of our methods. 

\subsection{Datasets and decision problems}
\label{subsec:datasets}

We consider four datasets, each
with an online decision problem:
\begin{itemize}
    \item \textbf{South Australian electricity market (AEMO).}
    We use half-hourly electricity price and demand data from 2022--2025 \citep{aemo2026pricedemand} to decide whether to curtail flexible load ahead of a price spike. Features include current and lagged prices and demand in South Australia and Victoria, as well as recent price maxima and demand changes. 
    The labels are \emph{normal price} and \emph{spike}, indicating whether the price exceeds $\$300$ per MWh in the half-hour interval ending one hour after the decision. The actions are \emph{run as scheduled} and \emph{curtail flexible load}.

    \item \textbf{Consumer complaints (CFPB).}
    We use complaints submitted to the Consumer Financial Protection Bureau during 2018--2024 \citep{cfpb2026complaints} to decide which incoming complaints to flag for review. Inputs include the consumer's narrative and metadata describing the product, issue, company, consumer's state,  and receipt date. The binary label records whether the company ultimately closes the complaint with monetary relief to the consumer. The actions are \emph{no review} and \emph{flag the complaint for review}.

    \item \textbf{Beijing air quality (Beijing).}
    We use hourly observations from 12 monitoring stations during 2013--2017 \citep{chen2019beijing} to choose an air-quality advisory for conditions six hours ahead. Features include station identity, pollutant and meteorological measurements, their lags and rolling averages, and calendar variables. Future PM$_{2.5}$ concentrations determine four ordered labels: \emph{good}, \emph{light pollution}, \emph{moderate pollution}, and \emph{heavy pollution}. The actions are \emph{none}, \emph{advise sensitive groups to reduce outdoor activity}, \emph{advise the public to reduce outdoor activity}, and \emph{emergency measures}.

    \item \textbf{Mississippi River flooding (Flood).}
    We use daily river-stage observations at St.~Louis during 1990--2025 \citep{usgs2026stlouis} to choose a protective action against flooding over the next three days. Features summarize current and recent river levels, changes in stage, rolling statistics, and seasonality. The highest daily stage over this horizon determines four ordered labels: \emph{below flood stage}, \emph{minor flood}, \emph{moderate flood}, and \emph{major flood}, following National Weather Service thresholds \citep{nws2026stlouis}. The actions are \emph{none}, \emph{protect low-lying areas}, \emph{partial evacuation}, and \emph{full evacuation}.
\end{itemize}

We process each series once in chronological order. The training period comprises the first $80\%$ of rounds on the AEMO and Flood datasets and the first three of four years on the Beijing dataset; the remaining observations form the test set. On these three datasets, a probabilistic classifier is fitted initially and refitted every $500$ test rounds using all time points whose labels have become available. On the CFPB dataset, Kev-4B \citep{palmer2026kev}, an open-source decision model fine-tuned from Qwen3.5-4B-Base \citep{qwen2026qwen35}, supplies probabilities from complaint text without further training. All methods receive the same probabilities and test observations. 
We provide details in Appendix~\ref{app:protocol}.

All four problems use the cost--loss model of forecast evaluation \citep{richardson2000skill}. Each outcome $y$ is assigned a harm $h_y$ reflecting its severity. We set $h=(0,1)$ on the AEMO and CFPB datasets. On the Beijing and Flood datasets, we assign harms using 
air-quality indices and flood depths (scaled appropriately), respectively, giving $h=(0,3,5,8)$ and 
$h=(0,5,15,30)$. 
The constructions are detailed in Appendix~\ref{app:datasets}.

Each action $a$ protects against harm up to $h_a$ at cost $fh_a$, giving total cost $fh_a+(h_y-h_a)_+$. The cost ratio $f\in(0,1)$ measures the cost per unit of protection relative to the harm prevented. Subtracting the minimum cost achievable for outcome $y$, namely $fh_y$, and dividing by $f$ gives
$$
\ell(a,y)
=(h_a-h_y)_+
+\frac{1-f}{f}(h_y-h_a)_+.
$$
Thus, each degree of over-protection costs one unit, whereas each degree of under-protection costs $(1-f)/f$ units.

This is normalized excess cost relative to the best action for the realized outcome. Writing $c(a,y)=fh_a+(h_y-h_a)_+$, we have $\sum_t c(a_t,y_t)=f\sum_t\ell(a_t,y_t)+f\sum_t h_{y_t}$, so cumulative rankings on a fixed stream are preserved. Percentage reductions below refer to $\ell$.
When $f<1/2$, insufficient protection is more costly than excess protection, 
more so as $f$ decreases.
We focus on $f=0.10$ and report $f\in\{0.05,0.25\}$ in Appendix~\ref{app:all-cost-ratios}. 

\subsection{Baselines and metrics}
\label{subsec:compared-methods}

We compare our approach
with a number of baselines. 
First we compare a number of  action-only baselines:
the best response (BR) method,
which applies the plug-in decision rule directly
on the predicted probabilities; Direct Loss Adaptation (DLA); and the Unbiased Prediction (UP) method of \citet{noarov2025high}, together with our variant UP-$\widehat p$. 
Among baselines that use prediction sets, we compare 
ACI \citep{gibbs2021adaptive} and DtACI \citep{gibbs2024conformal}, each combined with the robust action rule of \citet{wang2026optimal}.
 Additional details are provided below:

\begin{itemize}
    \item \textbf{Best response (BR)} uses 
    the best response \eqref{eq:best-response} to minimize plug-in expected loss, with ties broken toward the least protective action. 

    \item \textbf{Direct Loss Adaptation (DLA)} starts from the best response and corrects the predicted cost of each action by a discounted average of its past differences between the observed and predicted costs. 
    We use AdaHedge to combine the best response with four such corrected policies, which differ in how the residuals are pooled and discounted.
    The algorithm and an associated  regret bound are in Appendix~\ref{app:dla}.

    \item \textbf{Unbiased Prediction (UP)} \citep{noarov2025high} forecasts a distribution $q_t$ on $\Y$ from past labels only and takes the best response for this prediction,
    \eqref{eq:best-response} with $q_t$ in place of $\widehat p_t$. The forecasts are updated online to be unbiased conditional on the action this best response selects. Our variant, \textbf{UP with binned probabilities (UP-$\widehat p$)}, also conditions on equal-width bins of $\widehat p_t$, with the number of bins chosen on a validation period. Details are in Appendix~\ref{app:parameters}.

    \item \textbf{Ordinary ACI} maintains $\alpha_t$ by \eqref{eq:aci-update}, constructs $C_t=\{y:s_t(y)\leq q_t^0(\alpha_t)\}$, and chooses the robust action $a_t\in\argmin_a L(C_t,a;\alpha)$ of \eqref{eq:robust action} at the target level $\alpha$.

    \item \textbf{DtACI} maintains $\alpha_t$ using Algorithm~2 of \citet{gibbs2024conformal} (Appendix~\ref{app:parameters}). We use their step sizes and parameter values. The set is $C_t=\{y:s_t(y)\leq q_t^0(\alpha_t)\}$, and the action is the robust action at the target level $\alpha$, as for ordinary ACI.

    \item \textbf{DCP} uses the two surrogates $J_t$ and $V_t$ of Section~\ref{subsec:algorithm} in Algorithm~\ref{alg:dcp}: we run the algorithm with each surrogate and use AdaHedge to select between their actions, with weights determined by past decision losses.
This lets observed losses determine which criterion to favor. With immediate feedback, the cumulative loss, averaged over the action draws, exceeds that of the better of the two runs by at most $O(\sqrt T)$. With the delayed labels of our experiments, the weights are updated as the labels are observed (Appendix~\ref{app:delays}). The reported sets come from the run using $J_t$. 
We give the full procedure in Appendix~\ref{app:combination}
and report each run separately in Appendix~\ref{app:all-cost-ratios}.
\end{itemize}
 For all methods based on prediction sets, we 
use $\alpha=0.10$ and the scores $s_t(y)=1-\widehat p_t(y)$. 
Details
are given in Appendix~\ref{app:parameters}.

\noindent{\bf Metrics.}
Let $R$ denote the number of  rounds, let $n_r$ be the number of predictions in round $r$, and let $N_{\rm test}=\sum_{r=1}^R n_r$. 
We report the 
average coverage
$N_{\rm test}^{-1}\sum_{r,j}\mathbf1\{y_{r,j}\in C_{r,j}\}$, average decision loss $N_{\rm test}^{-1}\sum_{r,j}\ell(a_{r,j},y_{r,j})$, and average set size $N_{\rm test}^{-1}\sum_{r,j}|C_{r,j}|$. With one prediction per round, $N_{\rm test}=R=T$.

For DCP and DLA, we report the expected decision loss under their AdaHedge weights, and provide realized losses over $1{,}000$ independent draws of the actions in Appendix~\ref{app:realized}. 
For UP and UP-$\widehat p$, the 
selected
forecasts affect later updates, so we average the expected loss of each round over $20$ independent runs. 

We estimate standard errors by the stationary bootstrap \citep{politis1994stationary} of per-round outputs (Appendix~\ref{app:standard-errors}) and compare DCP with other methods using the paired $t$ statistic of their loss difference; negative values favor DCP.

\begin{table*}[!t]
  \centering
  \footnotesize
  \setlength{\tabcolsep}{8pt}
  \caption{\footnotesize Results at $f=0.10$ on the test set. Decision loss is reported as the mean $\pm$ the stationary-bootstrap standard error. The target coverage is $0.90$, and lower decision loss is better. Set size is the average number of labels in the set. BR, DLA, UP, and UP-$\widehat p$ form no set. Bold marks the lowest decision loss on a dataset when its paired $t$ statistics against all other methods are below $-1.96$.}
  \label{tab:main-results}
  \begin{tabular}{llccc}
    \toprule
    Dataset & Method & Coverage & Decision loss & Set size \\
    \midrule
      AEMO & BR & -- & $0.094$\se{0.014} & -- \\
       & DLA & -- & $0.061$\se{0.011} & -- \\
       & UP & -- & $0.124$\se{0.028} & -- \\
       & UP-$\widehat p$ & -- & $0.084$\se{0.018} & -- \\
       & Ordinary ACI & $0.905$ & $0.138$\se{0.012} & $0.94$ \\
       & DtACI & $0.909$ & $0.131$\se{0.010} & $0.94$ \\
       & DCP & $0.900$ & $\mathbf{0.055}$\se{0.010} & $0.92$ \\
      \addlinespace
      CFPB & BR & -- & $0.244$\se{0.012} & -- \\
       & DLA & -- & $0.207$\se{0.011} & -- \\
       & UP & -- & $0.298$\se{0.020} & -- \\
       & UP-$\widehat p$ & -- & $0.215$\se{0.014} & -- \\
       & Ordinary ACI & $0.903$ & $0.212$\se{0.015} & $0.92$ \\
       & DtACI & $0.904$ & $0.213$\se{0.015} & $0.92$ \\
       & DCP & $0.900$ & $\mathbf{0.202}$\se{0.011} & $0.97$ \\
      \addlinespace
      Beijing & BR & -- & $2.852$\se{0.096} & -- \\
       & DLA & -- & $2.823$\se{0.124} & -- \\
       & UP & -- & $5.850$\se{0.242} & -- \\
       & UP-$\widehat p$ & -- & $3.086$\se{0.136} & -- \\
       & Ordinary ACI & $0.901$ & $3.248$\se{0.154} & $1.80$ \\
       & DtACI & $0.902$ & $3.264$\se{0.150} & $1.82$ \\
       & DCP & $0.900$ & $2.749$\se{0.110} & $2.03$ \\
      \addlinespace
      Flood & BR & -- & $0.834$\se{0.371} & -- \\
       & DLA & -- & $0.823$\se{0.383} & -- \\
       & UP & -- & $3.061$\se{1.389} & -- \\
       & UP-$\widehat p$ & -- & $2.300$\se{1.222} & -- \\
       & Ordinary ACI & $0.912$ & $2.849$\se{0.624} & $0.94$ \\
       & DtACI & $0.905$ & $3.665$\se{0.535} & $1.00$ \\
       & DCP & $0.899$ & $0.769$\se{0.359} & $0.95$ \\
    \bottomrule
  \end{tabular}
\end{table*}

\subsection{Results}
\label{sec:results-headline}

At $f=0.10$, DCP has average coverage between $0.899$ and $0.900$, close to the target $0.90$, and has the lowest average decision loss in Table~\ref{tab:main-results}. Ordinary ACI and DtACI attain coverage between $0.901$ and $0.912$, but their decisions incur higher losses. DCP reduces loss by $5.0\%$ (CFPB) to $73.0\%$ (Flood) relative to ordinary ACI and by $5.3\%$ (CFPB) to $79.0\%$ (Flood) relative to DtACI. All eight paired $t$ statistics are below $-1.96$.

The improvement over the action-only baselines varies across datasets. On the AEMO and CFPB datasets, DCP reduces average loss by $42.1\%$ and $17.4\%$ relative to BR, by $10.4\%$ and $2.7\%$ relative to DLA, and by $35.3\%$ and $6.3\%$ relative to UP-$\widehat p$. All six paired $t$ statistics are below $-1.96$.
On the Beijing and Flood datasets, the reductions are $3.6\%$ and $7.8\%$ relative to BR, $2.6\%$ and $6.5\%$ relative to DLA, and $10.9\%$ and $66.6\%$ relative to UP-$\widehat p$; the paired $t$ statistics against BR on both datasets and against UP-$\widehat p$ on the Beijing dataset are below $-1.96$.
These results support the claim that our method has favorable performance compared to the baselines. 

\subsubsection{Visualizing the results}

\begin{figure}
\centering
\safeincludegraphics[width=0.8\textwidth]{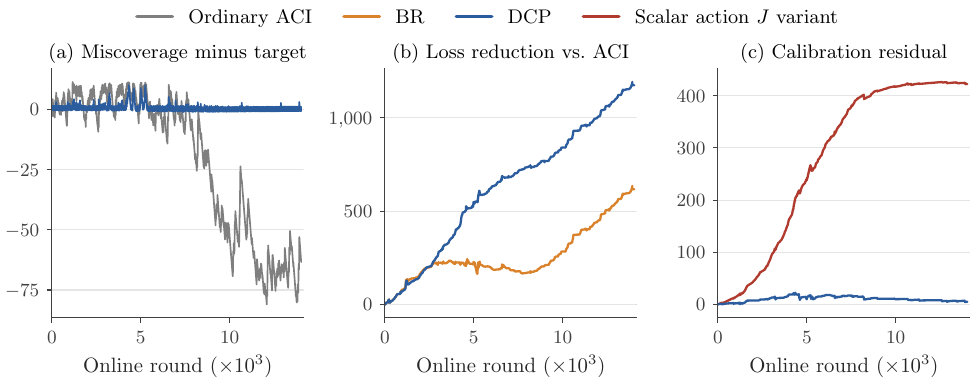}
\caption{\footnotesize A single run on the AEMO dataset at $f=0.10$. (a) Cumulative miscoverage minus the target, $\sum_{s\leq t}(e_s-\alpha)$, for ordinary ACI and DCP. (b) Cumulative decision loss reduction relative to ordinary ACI on the same run, $\sum_{s\leq t}\{\ell(a^{\rm ACI}_s,y_s)-\ell(a_s,y_s)\}$, for BR and DCP, with the loss of DCP taken as the expected loss under the AdaHedge weights. Positive values mean that the cumulative loss is lower than that of ordinary ACI. (c) Calibration of the out-of-set masses at the selected levels, $\sum_{y\in\Y}\big|\sum_{s\leq t}\{Z_{s,y}(\beta_s)-\widehat m^{\rm out}_{s,y}(\beta_s)\}\big|$ with $Z_{s,y}(\beta_s)=\1\{y_s=y,\ y_s\notin C_s\}$, for the calibrated masses of the run with $J_t$ and, computed along its own run, for the direct predicted masses $\1\{y\notin C_s\}\widehat p_s(y)$ of the scalar action $J$ variant of Appendix~\ref{app:uncalibrated}.}
\label{fig:trajectory-aemo}
\end{figure}

Figure~\ref{fig:trajectory-aemo} complements the average losses in Table~\ref{tab:main-results} by showing when the loss savings accumulate and how coverage and out-of-set mass estimates behave over time, for a run on the AEMO dataset at $f=0.10$.
Panel (a) shows that DCP keeps the cumulative number of misses within roughly 10 of its target throughout this run, while ordinary ACI eventually provides more coverage than required. Together with panel (b), this shows that DCP's lower decision loss is accompanied by near-target empirical coverage.

Panel (b) shows that DCP continues to accumulate loss savings relative to ordinary ACI during rounds $3{,}000$--$8{,}000$, when BR's savings largely stop growing.
At the same time, 
taking the best response to the predicted probabilities yields little additional savings.

Panel (c) examines the mass estimates used to choose actions. The cumulative discrepancy between corrected masses and observed out-of-set counts is bounded by
roughly $25$ over this trajectory.
In contrast, 
the discrepancy for direct predicted masses along the scalar action $J$ run reaches about $420$. 
This is consistent with feedback correction limiting accumulated errors in the quantities entering the robust action rule. Corresponding trajectories for the other datasets appear in Appendix~\ref{app:trajectories}.

\subsubsection{Diagnostics and ablations}

We now provide some basic diagnostics and ablations in order to better understand these results. 

{\bf A source of improvement: adverse event overestimation.}
First we aim to gain a deeper understanding of the source of the improvements over baselines.
For this,
one clue is that 
on both the AEMO and CFPB datasets,
the adverse event becomes rarer over time,
while the predictor overestimates its  frequency. 
For instance, on the AEMO dataset,
the average predicted probability is $4.2\%$,
while the observed frequency is $1.8\%$, 
despite refitting every $500$ rounds; a similar phenomenon is true for the CFPB dataset.
Since BR protects whenever the predicted probability exceeds $f$, this leads to unnecessary protective actions, which lead to a higher loss. 

In contrast, 
both DLA and DCP use label feedback to reduce protective actions. 
We observe that compared with DLA, DCP takes the protective action less often for observations whose predicted event probabilities lie just above $f=0.10$. 
In these ranges, the observed event frequency is below $10\%$.
For example, on the CFPB dataset, only $5.2\%$ of cases with predicted probabilities in $[0.10,0.15)$ experience an adverse event; DLA flags $18.9\%$ of these cases, compared with $3.2\%$ for DCP.
At $f=0.10$, each missed event costs nine times as much as a false alarm. 
Even after applying this weighting, DCP's reduction in false alarms more than offsets the cost of the additional misses on the CFPB dataset (Appendix~\ref{app:miscalibration}).
These patterns suggest
that DCP adapts better to the observed miscalibration.

In fact, in Appendix~\ref{subsec:dcp-improvement-br},
 we show that, under the present loss function, DCP can outperform BR in settings where adverse events are overestimated.
 We give a theoretical result and a numerical bound 
 computable based on observed quantities that can certify improvement.

{\bf The effect of the cost ratio.}
The performance of DCP relative to BR and DLA depends on the cost ratio. At $f=0.25$, BR and DLA have slightly lower average decision loss than DCP on the Beijing dataset (Appendix~\ref{app:all-cost-ratios}).
Under our normalized loss, each unit of over-protection costs $1$, whereas each unit of under-protection costs $(1-f)/f$. Thus, raising $f$ from $0.10$ to $0.25$ reduces the penalty for under-protection from $9$ to $3$, while leaving the penalty for over-protection unchanged.
DCP sometimes chooses a more protective action than BR. At the higher cost ratio, the benefit of this additional protection when it is needed is smaller relative to its cost when it is unnecessary, which can favor less protective decisions. This tradeoff is consistent with the small advantage of BR and DLA in this setting. 

{\bf Ablating the action.}
In Appendix~\ref{app:uncalibrated}, we report ablations of the action rule and the level selection. Replacing the corrected labelwise action rule \eqref{eq:corrected-action} by the
scalar robust rule \eqref{eq:robust action} increases the average loss in every evaluated setting. 
Moreover, 
using \eqref{eq:corrected-action} with the sets of ordinary ACI lowers the loss of ordinary ACI in all twelve settings.
This provides evidence that the labelwise action rule with corrected out-of-set masses is helpful in our experiments. 

{\bf Ablating the level.}
Running DCP with the single level $\alpha$ in place of the grid increases the loss on the Beijing dataset at every cost ratio and changes it by at most $3\%$ on the other datasets.

\section{Discussion}
\label{sec:discussion}

In this work we have developed methods that aim to control a decision loss in an online manner, targeted to settings where probability predictions are inaccurate, by leveraging prediction sets. 
We have argued that these methods can have empirical advantages over approaches that directly use predicted probabilities, such as best-response-type methods, when those predicted probabilities are inaccurate. 
A natural next step is to design the candidate sets themselves for the downstream loss, 
perhaps along the lines of the offline construction of \citet{wang2026optimal}.

\section*{Acknowledgements}
This work was supported in part by 
the ARO, ONR, and the Sloan Foundation. 

\setlength{\bibsep}{3pt plus 0.3ex}
\bibliographystyle{plainnat-abbrev}
\bibliography{ref}

\appendix

\section{Additional theoretical results and proofs}
\label{app:proofs}

\subsection{Proof of Proposition~\ref{thm:greedy-direct}}
\label{app:proof-greedy-direct}

\begin{proof}
For any comparator level, the selection rule gives
\begin{align*}
R_t(\beta_t)-R_t(\widetilde\beta_t)
&=J_t(\beta_t)-J_t(\widetilde\beta_t)
+\{R_t(\beta_t)-J_t(\beta_t)\}
-\{R_t(\widetilde\beta_t)-J_t(\widetilde\beta_t)\}
\le\omega_t.
\end{align*}
Conditional expectation converts this risk inequality into the
expected loss inequality. Summing proves \eqref{eq:greedy-candidate-regret}, and $\omega_t\le2\epsilon_t$ gives the bound $2\ell_{\max}\mathbb E\sum_{t=1}^T\epsilon_t$ under uniform validity.
\end{proof}

\subsection{Minimax interpretation of the DCP action}
\label{app:calibrated-action}

\begin{proposition}[Labelwise robust loss]
\label{prop:calibrated-action-minimax}
Let $C\subseteq\Y$ be nonempty and let $(m_y)_{y\notin C}$ satisfy $m_y\geq0$ and $\sum_{y\notin C}m_y\leq1$. Let $\mathcal U(C,m)$ be the set of distributions $Q$ on $\Y$ with $Q(y)\leq m_y$ for every $y\notin C$. Then, for every $a\in\A$,
\begin{equation}
\sup_{Q\in\mathcal U(C,m)}\E_{Y\sim Q}[\ell(a,Y)]
=
\ell_{\rm in}(C,a)+\sum_{y\notin C}m_y\big[\ell(a,y)-\ell_{\rm in}(C,a)\big]_+.
\label{eq:per-label-minimax}
\end{equation}
In particular, for nonempty $C_{t,\beta}$, the action \eqref{eq:corrected-action} is a minimizer of the worst-case expected loss over $\mathcal U(C_{t,\beta},\widehat m^{\rm out}_t(\beta))$.
\end{proposition}

\begin{proof}
Fix $a$ and write $l=\ell_{\rm in}(C,a)$. For every $Q\in\mathcal U(C,m)$, since $\ell(a,y)\leq l$ for $y\in C$ and $Q(C)=1-\sum_{y\notin C}Q(y)$,
$$
\E_{Y\sim Q}[\ell(a,Y)]
\leq Q(C)\,l+\sum_{y\notin C}Q(y)\ell(a,y)
=l+\sum_{y\notin C}Q(y)(\ell(a,y)-l)
\leq l+\sum_{y\notin C}m_y[\ell(a,y)-l]_+.
$$
To attain the bound, let $y^\star\in\argmax_{y\in C}\ell(a,y)$ and $S_a=\{y\notin C:\ell(a,y)>l\}$, and let $Q$ put mass $m_y$ on each $y\in S_a$, mass $1-\sum_{y\in S_a}m_y\geq0$ on $y^\star$, and no mass elsewhere. Then $Q\in\mathcal U(C,m)$ and $\E_{Y\sim Q}[\ell(a,Y)]$ equals the right-hand side of \eqref{eq:per-label-minimax}. The projection $\Pi_t$ makes $m_y=\widehat m^{\rm out}_{t,y}(\beta)$ nonnegative with sum at most one, which gives the last claim.
\end{proof}

\subsection{Plug-in sufficient condition for surrogate validity}
\label{app:proof-surrogate-validity}

\begin{proposition}[Plug-in sufficient condition for surrogate validity]
\label{prop:surrogate-validity}
Let $p_t^*(y)=\Prb(Y_t=y\mid\mathcal G_t)$ and use the normalized plug-in surrogate
\begin{equation}
J_t(\beta)=\sum_{y\in\Y}\widehat p_t(y)\bar\ell(a_{t,\beta},y).
\label{eq:plug-in-surrogate}
\end{equation}
Then
\begin{equation}
\max_{\beta\in\B}|J_t(\beta)-R_t(\beta)|
\leq\TV(\widehat p_t,p_t^*)
=\frac12\|\widehat p_t-p_t^*\|_1,
\label{eq:plugin-tv}
\end{equation}
and consequently $\omega_t\leq\|\widehat p_t-p_t^*\|_1$.
\end{proposition}
\begin{proof}
The candidate action is $\mathcal G_t$-measurable and both probability vectors sum to one. Centering the bounded loss therefore gives
\begin{align*}
|J_t(\beta)-R_t(\beta)|
&=\left|\sum_y(\widehat p_t(y)-p_t^*(y))
\left\{\bar\ell(a_{t,\beta},y)-\frac12\right\}\right|
\leq\frac12\|\widehat p_t-p_t^*\|_1.
\end{align*}
The bound is uniform over candidates. Taking the range of their errors gives the stated bound for $\omega_t$.
\end{proof}

\subsection{Proofs of Propositions~\ref{prop:greedy-plugin-bayes} and~\ref{prop:greedy-robust-bayes}}
\label{app:greedy-external}

We prove Propositions~\ref{prop:greedy-plugin-bayes}
and~\ref{prop:greedy-robust-bayes} for a general comparator action. Let
$a_t^\circ$ be any $\mathcal G_t$-measurable action. For example,
$a_t^\circ$ can be the action of another policy evaluated on the same
stream, if the covariate-label stream is exogenous to the deployed
actions and any randomization of that policy is included in
$\mathcal G_t$ and conditionally independent of $Y_t$ given the
information available before it. Let
$\Omega_t=\max_{a\in\mathcal A}\{r_t(a)-\widehat r_t(a)\}
-\min_{a\in\mathcal A}\{r_t(a)-\widehat r_t(a)\}$.

\begin{proposition}[Bounds against a general comparator action]
\label{prop:greedy-external}
For every such sequence $(a_t^\circ)$, the following hold.
\begin{enumerate}[label=(\roman*),leftmargin=2.2em,itemsep=0.1em]
\item For every round $t$ and candidate level $\beta$,
$r_t(a_{t,\beta})-r_t(a_t^\circ)\le\kappa_t(\beta,a_t^\circ)+\rho_t^-(\beta)+\rho_t^+(\beta)$.
Consequently, DCP with any surrogate satisfies
\begin{equation}
\mathbb E\sum_{t=1}^T
\{\ell(a_{t,\beta_t},Y_t)-\ell(a_t^\circ,Y_t)\}
\le\ell_{\max}\mathbb E\sum_{t=1}^T
\{\omega_t+[r_t^{\rm cand}-r_t(a_t^\circ)]_+\},
\label{eq:greedy-external-policy}
\end{equation}
and
$[r_t^{\rm cand}-r_t(a_t^\circ)]_+\le\min_{\beta\in\mathcal B}
\{\kappa_t(\beta,a_t^\circ)+\rho_t^-(\beta)+\rho_t^+(\beta)\}$.
\item At every round, DCP[$J$] satisfies
$r_t(a_{t,\beta_t})-r_t(a_t^\circ)\le\widehat\Gamma_t+\Omega_t$, and
$\Omega_t\le\|p_t^*-\widehat p_t\|_1$.
\item At every round, DCP[$V$] satisfies
$r_t(a_{t,\beta_t})-r_t(a_t^\circ)\le\rho_t^-(\beta_t)
+\min_{\beta\in\mathcal B}\{\kappa_t(\beta,a_t^\circ)+\rho_t^+(\beta)\}$.
\end{enumerate}
\end{proposition}

\begin{proof}
Write $a_t=a_{t,\beta_t}$,
$M_t(\beta,a)=\ell_{\rm in}(C_{t,\beta},a)/\ell_{\max}$, and
\begin{equation}
U_t(\beta,a)
=M_t(\beta,a)
+\sum_{y\notin C_{t,\beta}}\widehat m^{\rm out}_{t,y}(\beta)
[\bar\ell(a,y)-M_t(\beta,a)]_+
\label{eq:greedy-robust-objective}
\end{equation}
for the objective in \eqref{eq:corrected-action} divided by
$\ell_{\max}$. Thus
$a_{t,\beta}\in\operatorname*{argmin}_{a\in\mathcal A}U_t(\beta,a)$,
and the robust value is $V_t(\beta)=\min_{a\in\mathcal A}U_t(\beta,a)$.
Fix $\beta$ and $a$, and write $M=M_t(\beta,a)$. Since the probabilities
sum to one,
$r_t(a)=M+\sum_y p_t^*(y)\{\bar\ell(a,y)-M\}$.
Decompose each difference into its positive and negative parts.
The positive part is zero for every label in the candidate set, and the
negative parts give $-\kappa_t(\beta,a)$.
Subtracting this expression from \eqref{eq:greedy-robust-objective}
gives
\begin{equation}
U_t(\beta,a)-r_t(a)
=\kappa_t(\beta,a)
+\sum_{y\notin C_{t,\beta}}
\{\widehat m^{\rm out}_{t,y}(\beta)-p_t^*(y)\}
[\bar\ell(a,y)-M]_+.
\label{eq:greedy-robust-decomposition}
\end{equation}
For an empty set, $M=0$ and nonnegative losses make every negative part
zero, so $\kappa_t(\beta,a)=0$. Every positive-part loss difference in
\eqref{eq:greedy-robust-decomposition} lies in $[0,1]$ and
$\kappa_t(\beta,a)\ge0$. Therefore
\begin{equation}
r_t(a)-\rho_t^-(\beta)
\le U_t(\beta,a)
\le r_t(a)+\kappa_t(\beta,a)+\rho_t^+(\beta).
\label{eq:greedy-envelope-bounds}
\end{equation}

For (i), the robust minimizing property of a fixed candidate gives
\begin{align*}
r_t(a_{t,\beta})
&\le U_t(\beta,a_{t,\beta})+\rho_t^-(\beta)
\le U_t(\beta,a_t^\circ)+\rho_t^-(\beta)\\
&\le r_t(a_t^\circ)+\kappa_t(\beta,a_t^\circ)+\rho_t^-(\beta)+\rho_t^+(\beta).
\end{align*}
Minimizing over the candidates bounds $r_t^{\rm cand}-r_t(a_t^\circ)$,
and hence its positive part, since the right-hand side is nonnegative.
The proof of Proposition~\ref{thm:greedy-direct}, with $\widetilde\beta_t$
minimizing $R_t$, gives $R_t(\beta_t)-r_t^{\rm cand}\le\omega_t$, so
$R_t(\beta_t)-r_t(a_t^\circ)\le\omega_t+[r_t^{\rm cand}-r_t(a_t^\circ)]_+$.
Conditional expectation and summation prove
\eqref{eq:greedy-external-policy}.

For (ii), the selected action attains the predicted candidate minimum,
so
\begin{align*}
\widehat r_t(a_t)
&=\min_{a\in\mathcal A}\widehat r_t(a)+\widehat\Gamma_t
\le\widehat r_t(a_t^\circ)+\widehat\Gamma_t.
\end{align*}
Adding the two actions' forecast errors gives
$r_t(a_t)-r_t(a_t^\circ)\le\widehat\Gamma_t+\Omega_t$. For the
probability plug-in forecast, the facts that
$\sum_y(p_t^*(y)-\widehat p_t(y))=0$ and $0\le\bar\ell(a,y)\le1$ imply
\begin{align*}
|r_t(a)-\widehat r_t(a)|
&=\left|\sum_y(p_t^*(y)-\widehat p_t(y))
\{\bar\ell(a,y)-1/2\}\right|
\le\tfrac12\|p_t^*-\widehat p_t\|_1.
\end{align*}
Taking the range over actions gives
$\Omega_t\le\|p_t^*-\widehat p_t\|_1$.

For (iii), when $J_t=V_t$, the chosen pair $(\beta_t,a_t)$ jointly
minimizes $U_t(\beta,a)$ over the grid and action space. Applying
\eqref{eq:greedy-envelope-bounds} first at the chosen pair and then at
any pair $(\beta,a_t^\circ)$ gives
\begin{align*}
r_t(a_t)
&\le U_t(\beta_t,a_t)+\rho_t^-(\beta_t)
\le U_t(\beta,a_t^\circ)+\rho_t^-(\beta_t)\\
&\le r_t(a_t^\circ)+\kappa_t(\beta,a_t^\circ)
+\rho_t^+(\beta)+\rho_t^-(\beta_t).
\end{align*}
Minimizing the last expression over $\beta$ proves (iii).
\end{proof}

Multiplying (ii) and (iii) by $\ell_{\max}$, taking expectations, and
summing gives the corresponding regret bounds. Taking
$a_t^\circ=a_t^\star$, for which
$[r_t^{\rm cand}-r_t^\star]_+=\Gamma_t$, then gives
Corollary~\ref{cor:greedy-bayes} and
Propositions~\ref{prop:greedy-plugin-bayes}
and~\ref{prop:greedy-robust-bayes}.

The first bound in part (ii) also
holds when $\widehat r_t$ is replaced by other action-risk forecasts,
such as residual-corrected forecasts, with
$J_t(\beta)=\widehat r_t(a_{t,\beta})$ and $\widehat\Gamma_t$ computed
from these forecasts; if
$\max_{a\in\mathcal A}|r_t(a)-\widehat r_t(a)|\le\zeta_t$, then
$\Omega_t\le2\zeta_t$. In (iii),
conditional upper bounds on all out-of-set masses make
$\rho_t^-(\beta_t)=0$.
For an empty candidate set, $\kappa_t(\beta,a)=0$ for every action
$a$. More generally, if
$\ell_{\rm in}(C_{t,\beta},a_t^\star)\le\min_{y\in\Y}\ell(a_t^\star,y)+\ell_{\max}h_t$
for some $h_t\ge0$, then $\kappa_t(\beta,a_t^\star)\le h_t$.

\subsection{Necessity of the approximation terms}
\label{app:approximation-necessity}

Let the labels be i.i.d.\ on $\Y=\{1,2,3\}$ with probabilities
$(0.85,0.10,0.05)$ and no informative covariates. Take $\alpha=0.10$
and two actions $a_0,a_1$ with loss vectors $(0,2,0)$ and $(1,1,1)$,
respectively. Their expected losses are $0.20$ and $1$, so $a_0$ is
Bayes-optimal. A comparator policy $\pi$ always using $C_t^\pi=\{1,3\}$
has miscoverage probability exactly $\alpha$. With exact out-of-set
masses, its robust objective \eqref{eq:corrected-action} equals $0.20$
for $a_0$ and $1$ for $a_1$, so it selects $a_t^\pi=a_0$ at every round.

Run DCP with exact predicted probabilities and scores
$s_t(y)=1-\widehat p_t(y)$. Its threshold sets can only be
$\emptyset$, $\{1\}$, $\{1,2\}$, or $\{1,2,3\}$.
Whenever label $2$ is included, the robust objectives are $2$ and $1$
for $a_0$ and $a_1$, regardless of the corrected out-of-set masses.
Thus selecting $a_0$ requires $\emptyset$ or $\{1\}$, each with
miscoverage probability at least $0.15$.
Writing $N_0=\sum_{t=1}^T\mathbf 1\{a_t=a_0\}$, independence and the
expected coverage requirement give
$$
0.15\,\mathbb E N_0\le\mathbb E\sum_{t=1}^T e_t
\le0.10T+K_T.
$$
Here $K_T\ge0$ is the allowed coverage slack. Consequently,
\begin{equation}
\mathbb E\sum_{t=1}^T\{\ell(a_t,Y_t)-\ell(a_t^\pi,Y_t)\}
=0.80(T-\mathbb E N_0)
\ge\frac4{15}T-\frac{16}{3}K_T.
\label{eq:greedy-feasible-obstruction}
\end{equation}
For a fixed step size, \eqref{eq:greedy-direct-coverage} gives
$K_T=O(1)$, so this regret is linear in $T$.
For DCP[$J$], exact predicted probabilities also give $\omega_t=0$;
since $a_t^\pi$ is Bayes-optimal and $\ell_{\max}=2$,
Corollary~\ref{cor:greedy-bayes} implies
$
\mathbb E\sum_{t=1}^T\Gamma_t
\ge\frac2{15}T-\frac83K_T.
$
The accumulated approximation error therefore grows linearly even
against a comparator satisfying the same expected coverage requirement
and using the same robust action rule.

\subsection{Combining the plug-in surrogate and the robust value}
\label{app:combination}

The two surrogates offer different ways to compare candidate actions. 
With immediate feedback, we combine their actions using observed losses, obtaining regret of order $\sqrt T$ relative to the better run without assuming either surrogate is accurate. 

We run Algorithm~\ref{alg:dcp} twice on the same data stream, once with $J_t$ and once with $V_t$. Each run maintains its own $b_t$ and $d_t$ and produces its own set and action, $(C^J_t,a^J_t)$ and $(C^V_t,a^V_t)$. Both runs are updated after each observed label.

At round $t$, AdaHedge \citep{derooij2014follow} assigns weights $w_t=(w^J_t,w^V_t)$ to the two runs using their past losses $\ell(a^J_s,y_s)$ and $\ell(a^V_s,y_s)$, $s<t$. For $k\in\{J,V\}$, define $\lambda^k_t=\ell(a^k_t,y_t)/\ell_{\max}$, $L^k_t=\sum_{s=1}^t\lambda^k_s$, and $L^*_t=\min_{k\in\{J,V\}}L^k_t$. Initialize $L^k_0=0$ and $\Delta_0=0$. At round $t$, set $\eta_t=\ln 2/\Delta_{t-1}$ if $\Delta_{t-1}>0$ and $\eta_t=\infty$ otherwise. The weights in \eqref{eq:dcp-mixture-weights} are
\begin{equation}
w^k_t=
\begin{cases}
\displaystyle
\frac{\exp(-\eta_tL^k_{t-1})}
{\sum_{k'}\exp(-\eta_tL^{k'}_{t-1})},
& \eta_t<\infty,\\[1.2em]
\displaystyle
\frac{\1\{L^k_{t-1}=L^*_{t-1}\}}
{\sum_{k'}\1\{L^{k'}_{t-1}=L^*_{t-1}\}},
& \eta_t=\infty,
\end{cases}
\label{eq:adahedge-weights}
\end{equation}
where sums over $k'$ range over $\{J,V\}$. Thus, when $\eta_t=\infty$, the weights are uniform over the runs with smallest cumulative loss.

After observing $y_t$, compute $h_t=\sum_kw^k_t\lambda^k_t$ and
$$
m_t=
\begin{cases}
\displaystyle
-\frac{1}{\eta_t}\ln\sum_kw^k_te^{-\eta_t\lambda^k_t},
& \eta_t<\infty,\\[0.8em]
L^*_t-L^*_{t-1},
& \eta_t=\infty.
\end{cases}
$$
Update the cumulative mixability gap by $\Delta_t=\Delta_{t-1}+h_t-m_t$.

On the probability space introduced in Section~\ref{subsec:decision loss-regret}, the weights $w_t$ are $\mathcal F_{t-1}$-measurable. Let $\mathcal G_t$ be the sigma-algebra generated by $\mathcal F_{t-1}$ and the information of both runs before the action is drawn and $Y_t$ is observed. 
We choose $I_t\in\{J,V\}$ with fresh randomization satisfying $\mathbb P(I_t=k\mid\mathcal G_t,Y_t)=w^k_t$ for $k\in\{J,V\}$, 
and select $a^{I_t}_t$. 
Set $\mathcal F_t=\sigma(\mathcal G_t,I_t,Y_t)$.
The conditional expected loss of the chosen action is
$$
\bar\ell_t
=
w^J_t\,\ell(a^J_t,Y_t)
+
w^V_t\,\ell(a^V_t,Y_t)
=
\mathbb E\big[\ell(a^{I_t}_t,Y_t)\,\big|\,\mathcal G_t,Y_t\big].
$$
Here conditioning on $\mathcal G_t$ and $Y_t$ fixes the candidate actions, their weights, and the label; the expectation averages only over the fresh draw $I_t$. In \eqref{risks}, the label is instead left unconditioned and the action is fixed by $\mathcal G_t$.
The run with $J_t$ uses its own error $e^J_t=\1\{y_t\notin C^J_t\}$ in the calibration update, so \eqref{eq:greedy-direct-coverage} applies to these sets. When $I_t=V$, the chosen action is derived from $C^V_t$ and need not be the robust action associated with the reported set. 
Coverage therefore refers to the reported set sequence, 
while the following loss bound is a property of the combined actions.

\begin{proposition}[Regret of the combined procedure]
\label{prop:combination}
Let $R_{\rm AH}(T)=\ell_{\max}\{\sqrt{T\ln 2}+\tfrac43\ln 2+2\}$. Assume the immediate feedback protocol described above. For every realized data stream, with $\bar\ell_t$ evaluated at $Y_t=y_t$, the combined procedure satisfies
\begin{equation}
\sum_{t=1}^T\bar\ell_t
\leq
\min\Big\{\sum_{t=1}^T\ell(a^J_t,y_t),\ \sum_{t=1}^T\ell(a^V_t,y_t)\Big\}
+R_{\rm AH}(T).
\label{eq:combination-pathwise}
\end{equation}
Moreover, for every sequence of $\mathcal G_t$-measurable levels $\widetilde\beta_t\in\B$, with actions $a^J_{t,\widetilde\beta_t}$ of the run with $J_t$,
\begin{equation}
\mathbb E\sum_{t=1}^T\bar\ell_t
-
\mathbb E\sum_{t=1}^T\ell(a^J_{t,\widetilde\beta_t},Y_t)
\leq
\ell_{\max}\,\mathbb E\sum_{t=1}^T\omega^J_t
+R_{\rm AH}(T),
\label{eq:combination-expected}
\end{equation}
where $\omega^J_t$ is the surrogate error \eqref{eq:greedy-error-span} of the run with $J_t$.
\end{proposition}

Bound \eqref{eq:combination-expected} involves the surrogate error only of the run with $J_t$ and adds $R_{\rm AH}(T)$ to the bound of Proposition~\ref{thm:greedy-direct}.

\begin{proof}

Fix a realized data stream. The normalized expert losses $\lambda_t^k$ lie in $[0,1]$, and the loss of AdaHedge at round $t$ is $\bar\ell_t/\ell_{\max}$. Set $v_t=\max_k\lambda_t^k-\min_k\lambda_t^k$ and $v_{\max}=\max_{t\leq T}v_t$. Corollary~9 of \citet{derooij2014follow} gives
\begin{equation*}
\sum_{t=1}^T\frac{\bar\ell_t}{\ell_{\max}}
-\min_k\sum_{t=1}^T\lambda_t^k
\leq
\sqrt{\ln 2\sum_{t=1}^Tv_t^2}
+v_{\max}\left(\frac43\ln 2+2\right).
\end{equation*}
Since $v_t\leq1$ and $v_{\max}\leq1$, multiplication by $\ell_{\max}$ proves \eqref{eq:combination-pathwise}.

Taking expectations in \eqref{eq:combination-pathwise} and bounding the minimum by the loss of the run with $J_t$ gives $\mathbb E\sum_{t=1}^T\bar\ell_t\leq\mathbb E\sum_{t=1}^T\ell(a^J_t,Y_t)+R_{\rm AH}(T)$. 
Under the filtration defined above, the controllers of the run with $J_t$ remain $\mathcal F_{t-1}$-measurable, and its candidate actions and surrogate are $\mathcal G_t$-measurable. The condition $\mathbb P(I_t=k\mid\mathcal G_t,Y_t)=w^k_t$ also makes $I_t$ conditionally independent of $Y_t$ given $\mathcal G_t$.

The comparator levels are $\mathcal G_t$-measurable, and the regret bound of Proposition~\ref{thm:greedy-direct} uses only these measurability properties, with $R_t$ and $\omega^J_t$ defined given this $\mathcal G_t$. It therefore applies to the run with $J_t$ and gives
$$
\mathbb E\sum_{t=1}^T
\left\{
\ell(a^J_t,Y_t)
-
\ell(a^J_{t,\widetilde\beta_t},Y_t)
\right\}
\leq
\ell_{\max}\,
\mathbb E\sum_{t=1}^T\omega^J_t.
$$
Combining the two expected loss bounds proves \eqref{eq:combination-expected}.
\end{proof}

\paragraph{Sampled actions and a delayed feedback alternative.}
For immediate feedback and any fixed $T$ and $\delta\in(0,1)$, fresh conditionally independent action draws also give, with probability at least $1-\delta$,
\begin{equation}
\sum_{t=1}^T\ell(a_t^{I_t},Y_t)
\leq \min_{k\in\{J,V\}}\sum_{t=1}^T\ell(a_t^k,Y_t)
+R_{\rm AH}(T)
+\ell_{\max}\sqrt{\frac T2\log\frac1\delta}.
\label{eq:sampled-action-bound}
\end{equation}
To see this, condition on $(\mathcal G_t,Y_t)$. The sampled loss minus its conditional expectation $\bar\ell_t$ has conditional mean zero and range length at most $\ell_{\max}$. Its conditional exponential moment is at most $\exp(u^2\ell_{\max}^2/8)$ for parameter $u$. Iterating conditional expectations and optimizing the exponential Markov bound yields a tail probability at most $\exp\{-2z^2/(T\ell_{\max}^2)\}$ for a cumulative excess $z$. Combining with \eqref{eq:combination-pathwise} proves the claim.

For delayed labels, 
an alternative is to aggregate the two delayed DCP runs using the free copy construction in Appendix~\ref{subsec:dla-combination}, with two experts instead of five. In this construction, $t$ indexes individual predictions; simultaneous predictions may be ordered and assigned to different copies. Each copy is reused only after the losses of its previous prediction have arrived. If $Q_T$ copies are needed, its conditional expected loss satisfies
\begin{equation}
\sum_{t=1}^T\bar\ell_t^{\rm copy}
\leq \min_{k\in\{J,V\}}\sum_{t=1}^T\ell(a_t^k,y_t)
+\ell_{\max}\left\{\sqrt{Q_TT\ln2}
+Q_T\left(\frac43\ln2+2\right)\right\}.
\label{eq:dcp-delayed-copy}
\end{equation}
Indeed, the loss-vector sequence seen by each copy obeys the immediate feedback AdaHedge bound; summing these bounds and using $\sum_j\sqrt{n_j}\leq\sqrt{Q_TT}$ gives the result exactly as in Proposition~\ref{prop:dla-regret}. This argument allows the two underlying DCP runs to adapt using all arrived labels.
The same concentration argument applies conditional on a fixed exogenous stream and its fitted/controller paths, with fresh action draws independent of that stream and those paths.

\begingroup

\begingroup

\subsection{When DCP improves on the best response}
\label{subsec:dcp-improvement-br}

We now provide theoretical analysis
showing that DCP can have smaller decision
loss than BR in certain settings where the adverse event probabilities are overestimated.
Consider the binary cost--loss model of
Section~\ref{subsec:online-decisions}, with $\Y=\A=\{0,1\}$,
$\ell(0,1)=(1-f)/f$, $\ell(1,0)=1$, and zero diagonal losses.
Write
$p_t=p_t^*(1)=\mathbb P(Y_t=1\mid\mathcal G_t)$ for the conditional
event probability. 

BR takes a  protective action when $\widehat p_t(1)>f$, whereas
protection has smaller conditional expected loss only when $p_t>f$.
Thus, if $p_t<f<\widehat p_t(1)$, switching from protection to no
protection reduces conditional expected loss by
$(1-p_t)-(1-f)p_t/f=(f-p_t)/f>0$.
DCP can change this decision, 
and we now argue that this change can lead to 
an improvement in loss.
For the combined DCP procedure, let
$q_t=w_t^J\1\{a_t^J=1\}+w_t^V\1\{a_t^V=1\}$
be the probability of taking the protective action at prediction $t$.
Here $w_t^J$ and $w_t^V$ are the AdaHedge selection probabilities for DCP[$J$] and DCP[$V$], respectively, defined in \eqref{eq:dcp-mixture-weights}, and $a_t^J,a_t^V$ are the corresponding proposed actions.

The loss averaged over the fresh action draw is
$\bar\ell_t=(1-q_t)\ell(0,y_t)+q_t\ell(1,y_t)$.
 We will show an improvement in terms of the quantity 
\begin{equation}
\Phi_f(g,q,\nu)
=g-q-\frac{\min\{g,\nu\}-\max\{0,q+\nu-1\}}{f}.
\label{eq:dcp-br-frequency-function}
\end{equation}
 Our analysis will also depend on the following three terms, defined 
for any nonempty set $\mathcal I$ of $n$ binary observations: 
the fraction of protective actions taken by BR: 
$\bar g=n^{-1}\sum_{t\in\mathcal I}a_t^{\rm BR}$;
 the average probability assigned to protective actions by DCP: 
$\bar q=n^{-1}\sum_{t\in\mathcal I}q_t$; 
and the realized fraction of adverse events: 
$\nu=n^{-1}\sum_{t\in\mathcal I}y_t$.

To interpret $\Phi_f$, let
$\Delta_{\rm miss}=n^{-1}\sum_{t\in\mathcal I}
(a_t^{\rm BR}-q_t)y_t$ be DCP's additional  missed event rate
relative to BR, averaged under its action weights.
 We argue that 
 the exact loss reduction per
observation is $\bar g-\bar q-\Delta_{\rm miss}/f$.
Indeed, the term $\bar g-\bar q$ corresponds to a unit loss reduction per each avoided protective
action. 

When protection is removed on an adverse event, this reduction is reversed and the missed event cost is added, giving the factor
$1+(1-f)/f=1/f$.
The contribution of event observations to BR's overall protection
rate is $n^{-1}\sum_{t\in\mathcal I}a_t^{\rm BR}y_t$, normalized
by the total number $n$ of observations. This cannot exceed either
its overall protection rate $\bar g$ or the event frequency $\nu$,
since each protected event is both a protected observation and an
event. Hence this contribution is at most $\min\{\bar g,\nu\}$.

For DCP, nonevent observations make up a fraction $1-\nu$ of the
sample and each receives protection probability at most one.
They can therefore contribute at most $1-\nu$ to its overall
protection rate $\bar q$. The remaining contribution must come
from event observations:
\[
\frac1n\sum_{t\in\mathcal I}q_ty_t
=\bar q-\frac1n\sum_{t\in\mathcal I}q_t(1-y_t)
\ge\bar q-(1-\nu).
\]
Since this contribution is also nonnegative, it is at least
$\max\{0,\bar q+\nu-1\}$.
Thus, $\min\{\bar g,\nu\}-\max\{0,\bar q+\nu-1\}$ is the largest
value of $\Delta_{\rm miss}$ compatible with BR's protection rate
$\bar g$, DCP's protection rate $\bar q$, and the event frequency
$\nu$. This worst case occurs when BR protects on as many event
observations as possible, while DCP assigns as little protection
probability as possible to those observations.
Subtracting this quantity divided by
$f$ from $\bar g-\bar q$ gives $\Phi_f$.

We can formalize this intuition into the following bound about the certified improvement of DCP over BR in this model. 

\begin{proposition}[Certified improvement over BR]
\label{prop:dcp-br-frequency}

On every realized data stream, we have
\begin{equation}
\sum_{t\in\mathcal I}
\{\ell(a_t^{\rm BR},y_t)-\bar\ell_t\}
\ge n\Phi_f(\bar g,\bar q,\nu),
\label{eq:dcp-br-frequency-bound}
\end{equation}
which is a sharp bound in terms of $\bar g,\bar q,\nu$.
For any partition $(\mathcal I_i)_{i=1}^m$ of the test observations,
the total loss reduction is therefore at least
\begin{equation}
\mathcal C_T=\sum_{i=1}^m|\mathcal I_i|
\Phi_f(\bar g_i,\bar q_i,\nu_i).
\label{eq:dcp-br-certificate}
\end{equation}
\end{proposition}

In particular, consider a forecast band in which
$\widehat p_t(1)>f$ throughout, so that BR always protects and
$\bar g=1$. Then
\begin{equation}
\Phi_f(1,\bar q,\nu)
=1-\bar q-\frac{\min\{\nu,1-\bar q\}}{f},
\qquad
\Phi_f(1,\bar q,\nu)>0
\ \Longleftrightarrow\ \nu<f(1-\bar q).
\label{eq:dcp-br-overprotection-certificate}
\end{equation}
Here $1-\bar q$ is the fraction of protection removed, and
$\min\{\nu,1-\bar q\}$ is the largest fraction of observations
that could become missed events as a result.
Thus \emph{a low event frequency} combined with 
\emph{a sufficiently large
reduction in protection} certifies improvement even under the least
favorable association between DCP's actions and events.
The condition is stronger than $\nu<f$ because DCP could remove
protection mainly on event observations. 

{\bf Certified and empirical improvement.}
On the AEMO and CFPB datasets, Table~\ref{tab:miscalibration-bands}
shows forecast bands just above $f$ whose observed event frequencies
are below $f$, and in which DCP protects less often than BR.
This is consistent with the overestimation mechanism above.
For example, the displayed rates in the AEMO band $[0.10,0.15)$
give $\Phi_{0.10}(1,0.055,0.035)=0.595$ loss units per observation.

The resulting loss reduction depends on whether fewer false alarms
outweigh additional missed events. 
At $f=0.10$, each additional miss
costs nine false alarms. 
Table~\ref{tab:miscalibration-errors}
shows approximately $684$ fewer
false alarms and $14$ more
misses on the AEMO dataset, and $19{,}492$ fewer false alarms and
$1{,}205$ more misses on the CFPB dataset. These tradeoffs yield
reductions of approximately $684-9\times14=558$ and $19{,}492-9\times1{,}205=8{,}647$ in total loss.

Proposition~\ref{prop:dcp-br-frequency}, used 
for the five forecast
bands in Table~\ref{tab:miscalibration-bands}, implies the reductions in Table~\ref{tab:dcp-br-certificate}.
These amount to approximately $18.2\%$ and $6.5\%$
of their observed improvement, respectively.
The bound is conservative because it allows the least favorable association
between actions and outcomes within each forecast band.
The certified values allow a rounding error of up to $0.05$ percentage points in the event and DCP protection rates of each band in Table~\ref{tab:miscalibration-bands}, and in the BR protection rate of the $[0.10,0.15)$ band; the calculation is in Appendix~\ref{app:dcp-br-calculation}.

\begin{table}[H]
\centering
\footnotesize
\setlength{\tabcolsep}{6pt}
\caption{\footnotesize Certified and empirical reductions in DCP's
decision loss relative to BR at $f=0.10$, averaged under DCP's
action weights.}
\label{tab:dcp-br-certificate}
\begin{tabular}{lrrrr}
\toprule
& \multicolumn{2}{c}{Total reduction}
& \multicolumn{2}{c}{Mean reduction}\\
\cmidrule(lr){2-3}\cmidrule(lr){4-5}
Dataset & Certified & Empirical & Certified & Empirical\\
\midrule
AEMO & $101.5$ & $\approx558$
& $\approx0.007$ & $\approx0.040$\\
CFPB & $565.7$ & $\approx8{,}647$
& $\approx0.0028$ & $\approx0.043$\\
\bottomrule
\end{tabular}
\end{table}

\endgroup

\subsubsection{Proof of Proposition~\ref{prop:dcp-br-frequency}}
\label{app:dcp-improvement-br}

For binary outcomes, the loss can be written as
$\ell(a,y)=(1-f)y/f+a(1-y/f)$.
Consequently,
\begin{equation}
\frac1n\sum_{t\in\mathcal I}
\{\ell(a_t^{\rm BR},y_t)-\bar\ell_t\}
=\bar g-\bar q
-\frac1{nf}\sum_{t\in\mathcal I}(a_t^{\rm BR}-q_t)y_t.
\label{eq:dcp-br-binary-identity}
\end{equation}
Since $0\le a_t^{\rm BR},q_t,y_t\le1$,
\begin{align*}
\frac1n\sum_{t\in\mathcal I}a_t^{\rm BR}y_t
&\le\min\{\bar g,\nu\},\qquad
\frac1n\sum_{t\in\mathcal I}q_ty_t
\ge\max\{0,\bar q+\nu-1\}.
\end{align*}
For the second inequality, use both $q_ty_t\ge0$ and
$q_ty_t\ge q_t+y_t-1$.
Substitution in \eqref{eq:dcp-br-binary-identity} proves
\eqref{eq:dcp-br-frequency-bound}.

For sharpness, fix the event locations.
Assign BR's protective actions to event observations first, then
to nonevents if necessary. This attains
$n^{-1}\sum a_t^{\rm BR}y_t=\min\{\bar g,\nu\}$.
Independently assign DCP's protection probabilities to nonevents
first, then to events if necessary, using a fractional probability
on one observation if required. This attains
$n^{-1}\sum q_ty_t=\max\{0,\bar q+\nu-1\}$.
Both assignments are feasible simultaneously, subject to the
integer constraints already satisfied by the given empirical means.
Thus the bound cannot be increased using only those means.
Summing it over a partition proves the final claim.

\subsubsection{Additional empirical details}
\label{app:dcp-br-calculation}

{\bf Calculation from the reported diagnostic bands.}
At $f=0.10$, applying $\Phi_f$ to the printed frequencies in
Table~\ref{tab:miscalibration-bands} gives total lower bounds
$114.576$ for AEMO and $865.060$ for CFPB.
To account for rounding, allow each event frequency and DCP action
frequency to vary by $0.0005$ in either direction, clipped to $[0,1]$.

The BR threshold rule fixes its frequency at zero on $[0,0.10)$
and at one on every band starting at $0.15$ or above.
On $[0.10,0.15)$, allow its frequency to range over $[0.9995,1]$
for AEMO and $[0.9965,0.9975]$ for CFPB.
Over these ranges, the active branches of $\Phi_{0.10}$ do not change.
It equals $-\bar q$ on the bottom band,
$\bar g-\bar q-10\nu$ on bands with $\bar g>\nu$ and
$\bar q+\nu<1$, and $-9(1-\bar q)$ on bands with
$\bar g=1$ and $\bar q+\nu>1$.

Minimizing these linear expressions at their interval endpoints
gives the following lower bounds on each band's total contribution:
\begin{center}
\small
\begin{tabular}{lrr}
\toprule
Predicted probability & AEMO & CFPB\\
\midrule
$[0,0.10)$ & $-19.191$ & $-81.4435$\\
$[0.10,0.15)$ & $250.914$ & $8379.632$\\
$[0.15,0.20)$ & $72.8595$ & $-7668.315$\\
$[0.20,0.30)$ & $-175.269$ & $-33.0435$\\
$[0.30,1]$ & $-27.7875$ & $-31.0905$\\
\midrule
Total & $101.526$ & $565.7395$\\
\bottomrule
\end{tabular}
\end{center}

\endgroup

\section{Implementation and experimental details}
\label{app:protocol}

\subsection{Datasets and decision losses}
\label{app:datasets}

\paragraph{Datasets.}
Table~\ref{tab:dataset-setup} summarizes the four datasets. 
Here $N$ is the number of observations after preprocessing, $N_{\rm test}$ is the number in the test set, and the delay is the feedback lag used in the evaluation. A round can contain several observations. For Beijing the test set is the last full year, March 2016 to February 2017. For CFPB the input is the text of the complaint, and the test set runs from January 2019 to July 2024.

The CFPB complaints are taken from the Consumer Complaint Database \citep{cfpb2026complaints}, in a copy on Hugging Face made in January 2025 \citep{heineman2025cfpb}, because the database no longer publishes the consumers'
descriptions of the complaint in their own words. 
We use the complaints received from 1 July 2018 to 31 July 2024 that have a narrative and a final company response, and select $100$ of them at random on each day, or all complaints of a day when there are fewer. 

The database does not record when a complaint is closed. We therefore assume that the label of a complaint is observed $15$ days after it was sent to the company, the period within which companies generally respond, or $60$ days after for an untimely response. 
A timely response can be an interim one, with the final response due within $60$ days \citep{cfpb2026process}, so some labels may be used earlier than they were actually known.
The results with every label observed after $60$ days are in Appendix~\ref{app:cfpb-delay}.

\begin{table}[!htb]
  \centering
  \footnotesize
  \setlength{\tabcolsep}{4pt}
  \caption{\footnotesize Datasets after preprocessing.}
  \label{tab:dataset-setup}
  \begin{tabular}{llrrrl}
    \toprule
    Dataset & Round & $N$ & Features & $N_{\rm test}$ & Delay \\
    \midrule
    AEMO & 30 minutes, 2022--2025 & 70,120 & 27 & 14,024 & 1 hour \\
    CFPB & one day, 2018--2024 & 221,736 & text & 203,357 & 15 or 60 days (assumed) \\
    Beijing & one hour, 12 stations, 2013--2017 & 411,957 & 39 & 102,681 & 6 hours \\
    Flood & one day, 1990--2025 & 12,827 & 13 & 2,566 & 3 days \\
    \bottomrule
  \end{tabular}
\end{table}

\paragraph{Labels, actions, and decision losses.}
For all four datasets, the cost--loss model of Section~\ref{subsec:datasets} is
$$
\ell(a,y)=
\begin{cases}
h_a-h_y, & h_a\geq h_y,\\[2pt]
\dfrac{1-f}{f}\,(h_y-h_a), & h_a<h_y.
\end{cases}
$$
\begin{itemize}[leftmargin=1.8em,itemsep=0.15em]
\item \emph{AEMO.} The label is whether the regional price of South Australia in the half-hour interval that ends one hour after the decision exceeds $\$300$ per MWh, the strike price of the exchange-traded cap futures on this market \citep{asx2023electricity}. The actions are \emph{run as scheduled} and \emph{curtail flexible load}, with $h=(0,1)$.
Curtailing costs $f$ and an unprotected spike costs $1$.
\item \emph{CFPB.} The label is whether the company closes the complaint with monetary relief to the consumer; the other final responses are \emph{closed with explanation}, \emph{closed with non-monetary relief}, and \emph{untimely response}. The actions are \emph{no review} and \emph{flag the complaint for review}, with $h=(0,1)$: flagging costs $f$ and an unflagged complaint that is closed with monetary relief costs $1$.

\item \emph{Beijing.} The label is the PM$_{2.5}$ concentration six hours ahead, in ordinal concentration categories defined using the Chinese air quality index breakpoints \citep{mep2012aqi},
namely the 24-hour limits $75$, $115$, and $150~\mu\mathrm{g}/\mathrm{m}^3$: \emph{good} (excellent or good), \emph{light}, \emph{moderate}, and \emph{heavy} (heavy or severe) pollution. The 24-hour limits are applied to hourly concentrations. The actions are \emph{none}, \emph{advise sensitive groups to reduce outdoor activity}, \emph{advise the public to reduce outdoor activity}, and \emph{emergency measures}. We assign representative index values $50$, $125$, $175$, and $250$, subtract $50$, and divide by $25$, giving $h=(0,3,5,8)$. The value $250$ represents the merged heavy/severe category by the midpoint of the heavy-pollution interval $(200,300]$.

\item \emph{Flood.} The label is the highest daily stage over the next three days at the St.~Louis gauge \citep{usgs2026stlouis}, using the flood categories of the \citet{nws2026stlouis}: \emph{below flood stage} ($<30$ ft), \emph{minor} ($[30,35)$), \emph{moderate} ($[35,40)$), and \emph{major} ($\geq40$ ft). The actions are \emph{none}, \emph{protect low-lying areas}, \emph{partial evacuation}, and \emph{full evacuation}. The harm of a category is twice the depth in feet of its midpoint stage above flood stage, with the major category closed at the record crest of $49.58$ ft, which gives $h=(0,5,15,30)$ after rounding to integers.
\end{itemize}
At $f=0.10$, the matrices are
$$
\begin{gathered}
  \bigl(\ell^{\mathrm{Flood}}(a,y)\bigr)_{a,y}
  =
  \begin{pmatrix}
    0 & 45 & 135 & 270 \\
    5 & 0 & 90 & 225 \\
    15 & 10 & 0 & 135 \\
    30 & 25 & 15 & 0
  \end{pmatrix},
  \qquad
  \bigl(\ell^{\mathrm{BJ}}(a,y)\bigr)_{a,y}
  =
  \begin{pmatrix}
    0 & 27 & 45 & 72 \\
    3 & 0 & 18 & 45 \\
    5 & 2 & 0 & 27 \\
    8 & 5 & 3 & 0
  \end{pmatrix},
  \\[6pt]
  \bigl(\ell^{\mathrm{AEMO}}(a,y)\bigr)_{a,y}
  =
  \begin{pmatrix}
    0 & 9 \\
    1 & 0
  \end{pmatrix},
\end{gathered}
$$
with rows in action order and columns in label order; CFPB has the same matrix as AEMO. For other values of $f$, the entries below the diagonal are unchanged and those above it are multiplied by $(1-f)/(9f)$.

\subsection{Features and models}
\label{app:models}

\paragraph{Feature construction.}
For Flood, the features are the daily stage at 08:00, its values on the previous six days, its changes over one and three days, its seven-day maximum and thirty-day mean, and the sine and cosine of the day of the year. For Beijing, the features include station, current pollutant and meteorological variables, one-period lags, six-period rolling means, and calendar variables. For AEMO, five-minute prices and demands are averaged over half-hour intervals; the features are the price at lags zero to six and the demand at lags zero to two for South Australia and the neighbouring region of Victoria, the 24-hour maximum price and the one-hour change in demand for South Australia, and the time of day, day of the week, and day of the year. For CFPB, the text of a complaint consists of the date received, the product, sub-product, issue, and sub-issue, the company, the state of the consumer, and the channel of submission, followed by the consumer's narrative, cut to its first $6{,}000$ characters; all of these are known when the complaint is received.

\paragraph{Model fitting.}
On the AEMO, Beijing, and Flood datasets, numerical variables are median-imputed and standardized; categorical variables are mode-imputed and one-hot encoded. We fit multinomial logistic regression without class weights, using the L-BFGS solver and at most 500 iterations (2,000 for AEMO), on all timesteps whose labels have been observed, at the start of the test set and every $500$ rounds thereafter (Beijing: $500$ hours). The initial score history consists of the scores of the last $500$ timesteps of the training set (Beijing: $6{,}000$ rows) whose labels are observed at the start of the test set, computed with a classifier fitted on the timesteps labelled before these rows. DLA initializes its residual pools with the residuals of the same rows, computed with the same classifier.

\paragraph{Language model.}
On the CFPB dataset, $\widehat p_t$ are the probabilities of Kev-4B \citep{palmer2026kev}, revision \texttt{6cfce5c} of Kev~1.0, served as released with its fitted temperature. For each complaint it is asked one question with the two answers \emph{no: closed without monetary relief} and \emph{yes: closed with monetary relief}: 
``This consumer complaint was submitted to the CFPB and sent to the company. When the company responds, will it close the complaint with monetary relief, that is, a refund, credit, reimbursement or other payment to the consumer?'' The initial score history consists of the $16{,}749$ complaints received from July to December 2018 whose labels are observed by 1 January 2019, 
and DLA initializes its residual pools with the same complaints.

Kev-4B's model card at the specified revision reports a document-training stage using 5,219 CFPB narratives for product and main-issue questions.\footnote{\href{https://huggingface.co/jaredpalmer/kev-4b/blob/6cfce5c2fa4b4bd64026336ab649c5ca78857d52/README.md}{\textcolor{black}{Kev-4B model card, revision \texttt{6cfce5c}.}}} 
Thus, some of the evaluation data may have been present in the training corpus.
However, the questions of this stage concern the product and the issue of a complaint, not whether it is closed with monetary relief, and all methods use the same probabilities.

\subsection{Implementation of DCP}
\label{app:implementation}

The empirical quantile $q_t^0(\beta)$ in Section~\ref{sec:method} is the $\lceil(1-\beta)n\rceil$-th smallest of the $n$ scores in $H_{t-1}$.

The projection $\Pi_t$ of \eqref{eq:projection} is computed as follows. Order the grid as $\beta_1<\dots<\beta_K$ and collect the direct masses $\bar m^{\rm out}_{t,y}(\beta)$ into a $K\times|\Y|$ array. Four steps are applied in this order.
\begin{enumerate}[leftmargin=2.2em,itemsep=0.1em]
  \item \emph{Support.} Every entry with $y\in C_{t,\beta}$ is set to zero.
  \item \emph{Range.} The remaining entries are clipped to $[0,1]$.
  \item \emph{Monotonicity in $\beta$.} For each label separately, the entries along increasing $\beta$ are replaced by their isotonic regression, the least-squares nondecreasing fit, computed by pooling adjacent violators.
  \item \emph{Sub-probability.} Let $M_t$ be the largest row sum after step~3. If $M_t>1$, every entry of the array is multiplied by $1/M_t$; otherwise the array is unchanged.
\end{enumerate}
Since the candidate sets are nested, the entries set to zero in step~1 come first along increasing $\beta$, so step~3 keeps them at zero; it also keeps all entries in $[0,1]$. The factor in step~4 is shared across all $\beta$, so it preserves the zeros of step~1 and the ordering of step~3, and the output satisfies the support, sub-probability, and monotonicity constraints of Section~\ref{subsec:surrogate}.

\subsection{Compared methods and parameters}
\label{app:parameters}

\paragraph{DtACI.}
DtACI \citep[Algorithm~2]{gibbs2024conformal} keeps $k$ levels $\alpha_{t,i}$ with step sizes $\gamma_i$ and weights $w_{t,i}$, and uses the level $\alpha_t=\sum_iw_{t,i}\alpha_{t,i}/\sum_iw_{t,i}$. Let $C_t(\theta)=\{y:s_t(y)\leq q_t^0(\theta)\}$ and $\theta^\star_t=\sup\{\theta:y_t\in C_t(\theta)\}$. When $y_t$ is observed, set $\bar w_{t,i}=w_{t,i}\exp\{-\eta\,\psi(\theta^\star_t,\alpha_{t,i})\}$ and update the weights and levels by
$$
w_{t+1,i}=(1-\sigma)\bar w_{t,i}+\frac{\sigma}{k}\sum_{j=1}^k\bar w_{t,j},
\qquad
\alpha_{t+1,i}=\alpha_{t,i}+\gamma_i\big(\alpha-\1\{y_t\notin C_t(\alpha_{t,i})\}\big),
$$
where $\psi(\theta',\theta)=\alpha(\theta'-\theta)-\min\{0,\theta'-\theta\}$ is the pinball loss, and $C_t(\cdot)$ and $\theta^\star_t$ are computed from the score history at round $t$.

\paragraph{Unbiased Prediction.}
\citet{noarov2025high} do not provide an implementation, so we implement UP from their paper. The label of round $t$ is encoded as $s_t=e_{y_t}\in\Delta(\Y)$. As in their Algorithm~1, each event $E$, label $y$, and sign $\sigma\in\{-1,1\}$ define an expert with loss $\sigma E(q_t)\{s_t(y)-q_t(y)\}$, and MsMwC \citep{chen2021impossible},
with the parameters of its Theorem~2 and the horizon bound $10^6$, assigns weights to these experts. 

At each round, the forecast $q_t$ is drawn from the solution of the minimax problem of \citet[Section~2]{noarov2025high}, and the action is the best response to $q_t$, with ties broken as for BR. UP uses the events $\{\mathrm{BR}(q_t)=a\}$, $a\in\A$. UP-$\widehat p$ uses the products of these events with the bins $[0,1/G),\ldots,[(G-1)/G,1]$ of $\widehat p_t(y)$ for each label $y$, or of the event probability only in the binary problems.

In both versions, the events that occur in a round depend on $q_t$ only through its best response, so 
we use a linear programming formulation with one candidate forecast associated with each best-response region, as in \citet[Appendix~D]{noarov2025high}. Each candidate forecast is moved slightly into the interior of its region, by a convex combination with weight $10^{-6}$ on an interior point, so that its best response is the action of that region regardless of the rule for ties. The number of bins $G\in\{2,5,10\}$ is chosen separately for each dataset and cost ratio by the average decision loss on a validation period, the last $20\%$ of the training period (on the CFPB dataset, of the complaints before the test set), with the predicted probabilities computed as in the test period.

At $f=0.05$, $0.10$, and $0.25$, the chosen $G$ is $10$, $10$, and $10$ on the AEMO dataset, $10$, $5$, and $2$ on the CFPB dataset, $10$, $10$, and $10$ on the Beijing dataset, and $10$, $5$, and $5$ on the Flood dataset. Both versions are first run over the rows of the initial score history with their labels. The reported loss is the expected loss under the distribution of $q_t$, averaged over $20$ runs with independent draws. The Monte Carlo standard error of this average, the standard deviation over the runs divided by $\sqrt{20}$, is at most $9\%$ of its bootstrap standard error for UP and at most $4\%$ for UP-$\widehat p$ in all twelve settings.

\paragraph{Parameters.}
Across all experiments the base parameters are $\gamma=0.01$, $\eta_b=0.05$, $\eta_m=0.02$, and $c_{\max}=0.25$, with grid $\B=\{0.02,0.03,\ldots,0.30\}$. DtACI uses the $k=8$ step sizes $\gamma_i=0.001\cdot2^{i-1}$, $i=1,\ldots,8$, of \citet{gibbs2024conformal} and starts from $\alpha_{1,i}=\alpha$ and equal weights, with their choices of $\sigma$ and $\eta$ for an interval of length $500$, $\sigma=1/1000$ and $\eta=[3\{\log(500k)+2\}/\{500(1-\alpha)^2\alpha^2\}]^{1/2}$. DLA uses the half-lives $50$ and $200$ and the shrinkage $\kappa=10$.

\subsection{Label delays and several predictions per round}
\label{app:delays}

\paragraph{Delayed labels.}
In all four problems, the label of a round is observed later: $D=2$ rounds later on the AEMO dataset (1 hour), $D=3$ rounds later on the Flood dataset (3 days, or later where days are missing from the gauge record), $D=6$ rounds later on the Beijing dataset (6 hours), and on the CFPB dataset $15$ days after the complaint was sent to the company, or $60$ days for an untimely response (Appendix~\ref{app:datasets}). Each method forms the set and the action of a round before its label is observed. DCP stores the candidate sets, the projected masses $\widehat m^{\rm out}_t(\beta)$, the level $\beta_t$, and the scores $s_t(\cdot)$. The updates of $d_t$ and $b_t$, and the appending of $s_t(y_t)$ to the score history, are applied at the first round at which $y_t$ has been observed, with the stored quantities; labels observed at the same round are processed in the order of their rounds. Ordinary ACI and DtACI update their levels in the same way, and UP and UP-$\widehat p$ add the gains of a round to their expert weights at the first round at which its label has been observed, in the same order.

Delayed feedback can increase controller overshoot because several predictions can be issued before the effect of an update is observed. As a tuning heuristic, we cap the step sizes of the updates that use labels at $\min(\gamma,0.5/D)$, $\min(\eta_b,0.5/D)$, and $\min(\eta_m,0.5/D)$, also for the scalar action variants of Appendix~\ref{app:uncalibrated}, with $D=15$ on the CFPB dataset.
 For DCP and ordinary ACI, the cap changes only $\eta_b$ on the CFPB dataset, to $1/30$. DtACI caps each of its step sizes $\gamma_i$ at $0.5/D$ in the same way, which lowers the largest to $1/12$ on the Beijing dataset and the two largest to $1/30$ on the CFPB dataset.

The AdaHedge combination of the two runs of DCP (Appendix~\ref{app:combination}) adds the losses of each prediction to the weights when its label is observed, in the order of arrival. This is the 
implementation evaluated in the experiments; Proposition~\ref{prop:combination} establishes the stated regret bound for immediate feedback. DLA uses copies of AdaHedge to handle delays, with a
corresponding guarantee given in Appendix~\ref{subsec:dla-regret}.

\paragraph{Several predictions per round.}
On the Beijing and CFPB datasets, one round contains several predictions: the twelve stations of an hour, or the complaints received on a day. All predictions of a round use the same controller state. The labels of one round that are observed at the same later round enter one update of $d_t$ and $b_t$ with the mean increment over these labels. On the Beijing dataset, the labels of an hour are observed together. On the CFPB dataset, the labels of the complaints of one day arrive on different days; each update then multiplies $\eta_b$ and $\eta_m$ by the fraction of the day's complaints whose labels it contains, so that the labels of one day together give one full step.

Ordinary ACI and DtACI are run in the same way: with several labels in one update, the errors, and for DtACI the pinball losses, are averaged over the labels, and on the CFPB dataset $\gamma$, the $\gamma_i$, and $\eta$ are multiplied by the same fraction. DLA, UP, and UP-$\widehat p$ treat the predictions of one round as consecutive rounds, each with the arrival time of its own label.

\paragraph{Ties with delayed labels.}
With delayed labels, $b_t$ does not yet contain the errors of the predictions whose labels are unobserved. The rule for ties in Algorithm~\ref{alg:dcp} therefore uses
$$
\widehat b_t=b_t+\eta_b\sum_{s\in P_t}w_s(\beta_s-\alpha)
$$
in place of $b_t$, where $P_t$ is the set of earlier predictions whose labels have not entered $b_t$, including the earlier predictions of the same round, $\beta_s$ is the level selected for prediction $s$, and $w_s$ is its weight in the update of $b_t$: $1$, or $1/n$ for a round with $n$ predictions. Thus each unobserved error $e_s$ is replaced by $\beta_s$. Among the levels minimizing $J_t$, DCP takes the largest if $\widehat b_t<0$ and the smallest otherwise. The variable $\widehat b_t$ is used only for this choice, and the update of $b_t$ is unchanged. With immediate feedback and one prediction per round, $P_t$ is empty and $\widehat b_t=b_t$.

\begin{proposition}[Coverage with delayed and grouped feedback]
\label{prop:delayed-coverage}
In prediction round $r$, issue finitely many predictions indexed by $j$,
with weights $w_{r,j}\ge0$ fixed at issuance and satisfying
$\sum_jw_{r,j}=1$. Write
$e_{r,j}=\mathbf1\{y_{r,j}\notin C_{r,j}\}$ and
$\bar e_r=\sum_jw_{r,j}e_{r,j}$.
Suppose the scores and the unshifted thresholds are in $[0,1]$.
Initialize $b=0$. When a label is processed, add
$\eta_bw_{r,j}(e_{r,j}-\alpha)$ to $b$, using the stored prediction
set. 
Suppose that the total weight of issued predictions whose increments
remain unprocessed never exceeds $P$, including immediately after
a new prediction batch is issued. Then every controller state satisfies
$$
-1-\eta_b\alpha P\le b\le1+\eta_b(1-\alpha)P.
$$
After the first $R$ prediction rounds, take an endpoint after processing
any chosen arrivals and before issuing round $R+1$. If $W_R\le P$ is
the total weight still pending at this endpoint, then
\begin{equation}
\left|\frac1R\sum_{r=1}^R\bar e_r-\alpha\right|
\le\frac{\eta_b^{-1}+\max\{\alpha,1-\alpha\}(P+W_R)}{R}
\le\frac{\eta_b^{-1}+2P\max\{\alpha,1-\alpha\}}{R}.
\label{eq:delayed-weighted-coverage}
\end{equation}
\end{proposition}

\begin{proof}
For the analysis of $b$, expand each grouped update into its weighted increments. 
Between issuance events, pending weight can only decrease, so its
bound $P$ also holds immediately before each atomic update.

Consider an excursion above one and the update that starts it from a
state at most one. Every prediction issued subsequently while the
state remains above one has a full prediction set, hence $e=0$ and a
nonpositive eventual update. All positive increments during this
excursion therefore belong to predictions already pending immediately
before the crossing, including the increment that causes the crossing.
Their total weight is at most $P$, and their total positive contribution
is at most $\eta_b(1-\alpha)P$. This proves
$b\le1+\eta_b(1-\alpha)P$. During an excursion below minus one,
every newly issued prediction has an empty set and hence a nonnegative
eventual update. Only the backlog present immediately before its
downcrossing can decrease the state, by at most $\eta_b\alpha P$ in
total. Thus $b\ge-1-\eta_b\alpha P$. The initial state lies between
both barriers, proving the invariant bounds.

Let $b_R^{\rm obs}$ be the endpoint state and let $\mathcal P_R$
index the still-unprocessed predictions from rounds $1,\ldots,R$.
Summing the processed increments gives
\begin{equation}
\sum_{r=1}^R(\bar e_r-\alpha)
=\frac{b_R^{\rm obs}}{\eta_b}
+\sum_{(r,j)\in\mathcal P_R}w_{r,j}(e_{r,j}-\alpha).
\label{eq:delayed-coverage-accounting}
\end{equation}
The last sum lies in $[-\alpha W_R,(1-\alpha)W_R]$.
Together with the invariant interval, this bounds the right-hand side
between $-\eta_b^{-1}-\alpha(P+W_R)$ and
$\eta_b^{-1}+(1-\alpha)(P+W_R)$. Taking absolute values and
dividing by $R$ proves \eqref{eq:delayed-weighted-coverage}.
\end{proof}

\subsection{Standard errors}
\label{app:standard-errors}

We apply the stationary bootstrap \citep{politis1994stationary} to the per-round outputs of each method, with circular indexing and restart probability $0.01$ (geometric blocks with mean length $100$), and $B=1{,}000$ replicates; the standard error is the standard deviation of the $B$ replicate means. The paired $t$ statistic of two methods is the mean of the per-round differences of their losses divided by its standard error from the same bootstrap. For Beijing the bootstrap resamples hours, with all stations of an hour together, and for CFPB it resamples days, with all complaints of a day together.

\section{Direct Loss Adaptation}
\label{app:dla}

\emph{Direct Loss Adaptation} (DLA) is the action-only adaptive baseline in Section~\ref{sec:experiments}. The best response can incur excess loss when the predicted probabilities give inaccurate action costs; DLA addresses this problem directly, without imposing a coverage target or forming prediction sets. It uses the same probabilities and observed labels as DCP, corrects predicted action costs by weighted averages of past cost residuals, and combines several corrected policies with the original best response using AdaHedge. 

\subsection{Corrected action costs}
\label{subsec:dla-costs}

For each action $a\in\A$, the predicted normalized cost is
\begin{equation}
r^0_t(a)=\frac{1}{\ell_{\max}}\sum_{y\in\Y}\widehat p_t(y)\ell(a,y).
\label{eq:dla-base-risk}
\end{equation}
The best response chooses $g_t\in\argmin_{a\in\A}r^0_t(a)$, with ties broken toward the least protective action. When $y_s$ is observed, the loss matrix gives the cost of every action, not only of the action taken, and the cost residual is
\begin{equation}
\xi_s(a)=\frac{\ell(a,y_s)}{\ell_{\max}}-r^0_s(a),
\qquad a\in\A.
\label{eq:dla-residual}
\end{equation}
If the predicted probabilities understate the cost of action $a$, its residuals are positive on average; if they overstate it, the residuals are negative on average.

DLA estimates the mean residual by discounted least squares. 
A set of past residuals
 stores discounted sums $S(a)$, $a\in\A$, and $N$, all initialized at zero, and updates them for each residual it receives by
\begin{equation}
S(a)\leftarrow\rho_\tau S(a)+\xi_s(a),
\qquad
N\leftarrow\rho_\tau N+1,
\qquad
\rho_\tau=2^{-1/\tau}.
\label{eq:dla-residual-update}
\end{equation}
After $n$ residuals $\xi_1,\ldots,\xi_n$, the estimate $S(a)/(\kappa+N)$ minimizes $\kappa u^2+\sum_{j=1}^n\rho_\tau^{n-j}\{\xi_j(a)-u\}^2$ over $u\in\R$, so the shrinkage $\kappa>0$ keeps it near zero until enough residuals have been received (Appendix~\ref{app:dla-least-squares}). A \emph{global} pool receives every residual. A \emph{context} pool receives only the residuals of rounds with the same best-response action $g_s$, so that the correction may differ between rounds in which the classifier recommends different actions. The corrected policy of pool $k$ chooses
\begin{equation}
a^k_t\in\argmin_{a\in\A}\widehat r^k_t(a),
\qquad
\widehat r^k_t(a)=\clip_{[0,1]}\left(r^0_t(a)+\frac{S^k_t(a)}{\kappa+N^k_t}\right),
\label{eq:dla-corrected-policy}
\end{equation}
with ties broken as for $g_t$, where $S^k_t$ and $N^k_t$ are the sums of the pool before round $t$; for a context pool they are read from the pool of the current $g_t$. The policy keeps the information about the current round in $r^0_t$ and shifts each predicted cost by a correction learned from past rounds. The global and context pools with half-lives $\tau\in\{50,200\}$, together with the best response, give $M=5$ policies, called experts,
$$
\mathcal E=\{0,\ (\mathrm{global},50),\ (\mathrm{global},200),\ (\mathrm{context},50),\ (\mathrm{context},200)\},
$$
where expert zero is the best response, $a^0_t=g_t$ and $\widehat r^0_t=r^0_t$.

\subsection{Combining the experts}
\label{subsec:dla-combination}

At round $t$, AdaHedge assigns weights $w^k_t$, $k\in\mathcal E$, to the experts from their normalized losses $\lambda^k_s=\ell(a^k_s,y_s)/\ell_{\max}$ on the rounds whose labels have been observed, by the update of Appendix~\ref{app:combination} with $\ln 2$ replaced by $\ln M$. DLA draws $I_t\in\mathcal E$ with probabilities $w_t$, using fresh randomization, and deploys $a^{I_t}_t$. On the probability space of Section~\ref{subsec:decision loss-regret}, let $\mathcal G_t$ be the sigma-algebra generated by the information available before $I_t$ is drawn and $Y_t$ is observed, including the covariates, predicted probabilities, labels observed so far, expert actions, and states of the pools and of AdaHedge. Fresh randomization means $\mathbb P(I_t=k\mid\mathcal G_t,Y_t)=w^k_t$. As in Appendix~\ref{app:combination}, the conditional expected loss of the chosen action is
$$
\bar\ell^{\rm DLA}_t=\sum_{k\in\mathcal E}w^k_t\ell(a^k_t,Y_t)=\E\big[\ell(a^{I_t}_t,Y_t)\,\big|\,\mathcal G_t,Y_t\big].
$$
This expectation averages over the expert draw $I_t$ with the label and expert actions fixed by the conditioning. The conditional risks $r_t^*(a)$ used below instead average over $Y_t$ given $\mathcal G_t$, as in \eqref{risks}; with delayed feedback, this information contains only labels that have arrived.
In our experiments, labels are observed with a delay (Appendix~\ref{app:delays}), so the losses of round $t$ may not be available before round $t+1$. DLA handles this by the reduction of \citet{joulani2013delayed}. It keeps several copies of AdaHedge, each with its own cumulative normalized losses $L^k$, $k\in\mathcal E$, and mixability gap $\Delta$. A copy is busy from the round at which it is selected until the label of that round arrives, and free otherwise. 

At each round, after the arrived labels have been processed, DLA selects the free copy with the smallest index, or creates a new copy with $L^k=0$ and $\Delta=0$ if all copies are busy. Each round stores $r^0_t$, $g_t$, the actions $a^k_t$ of all experts, and the selected copy. When $y_t$ arrives, that copy is updated with the losses $\lambda^k_t$ of the stored actions, and the residual pools, which all copies share, are updated with $\xi_t$ computed from the stored $r^0_t$ and $g_t$, so that a later refit of the classifier leaves it unchanged. Labels that arrive at the same time are processed in the order of their rounds.
Algorithm~\ref{alg:dla} summarizes the procedure.

\begin{algorithm}[t]
\caption{\footnotesize Direct Loss Adaptation (DLA)}
\label{alg:dla}
\begin{algorithmic}[1]
\REQUIRE loss function $\ell$, half-lives $\{50,200\}$, shrinkage $\kappa>0$, and optional labelled rounds before the first round.
\STATE Initialize all residual pools at zero and update them with the residuals of the labelled initial rounds, in time order, by \eqref{eq:dla-residual-update}. Create one copy of AdaHedge with zero cumulative losses.
\FOR{$t=1,2,\ldots,T$}
\FOR{each label $y_s$ that has arrived and has not been processed, in order of arrival}
\STATE Compute $\xi_s$ by \eqref{eq:dla-residual} with the stored $r^0_s$, and update the two global pools and the two context pools of $g_s$ by \eqref{eq:dla-residual-update}.
\STATE Update the copy of AdaHedge used at round $s$ with the losses $\lambda^k_s$ of the stored actions $a^k_s$, $k\in\mathcal E$.
\ENDFOR
\STATE Observe $x_t$ and compute $\widehat p_t$, $r^0_t$, $g_t$, and $a^k_t$ for $k\in\mathcal E$ by \eqref{eq:dla-base-risk} and \eqref{eq:dla-corrected-policy}.
\STATE Select the first copy of AdaHedge whose previous round has been labelled, or create a new copy; let $w_t$ be its weights.
\STATE Select $I_t$ with probabilities $w_t$, deploy $a_t=a^{I_t}_t$, and store $r^0_t$, $g_t$, $(a^k_t)_{k\in\mathcal E}$, and the selected copy.
\ENDFOR
\end{algorithmic}
\end{algorithm}

\subsection{Regret guarantee}
\label{subsec:dla-regret}

Let $Q_T$ be the number of copies of AdaHedge created up to round $T$. A new copy is created only when every existing copy is waiting for a label, so $Q_T$ is at most one plus the largest number of unobserved labels at a round. In particular, $Q_T=1$ when each label is observed before the next round.

\begin{proposition}[Regret of DLA]
\label{prop:dla-regret}
Let $c_M=\tfrac43\ln M+2$. For every realized data stream and all label arrival times, with $\bar\ell^{\rm DLA}_t$ evaluated at $Y_t=y_t$, Algorithm~\ref{alg:dla} satisfies
\begin{equation}
\sum_{t=1}^T\bar\ell^{\rm DLA}_t
\leq
\min_{k\in\mathcal E}\sum_{t=1}^T\ell(a^k_t,y_t)
+\ell_{\max}\left\{\sqrt{Q_TT\ln M}+Q_Tc_M\right\}.
\label{eq:dla-regret}
\end{equation}
\end{proposition}

The proof is in Appendix~\ref{app:proof-dla-regret}. The comparator is the best single expert over the whole horizon, evaluated along the run of DLA. It includes the best response, $k=0$, so the cumulative loss of DLA exceeds that of BR by at most the second term of \eqref{eq:dla-regret}. With $Q_T=1$, the second term is $R_{\rm AH}(T)$ of Proposition~\ref{prop:combination} with $\ln 2$ replaced by $\ln M$. The copies extend the bound to delayed labels at the cost of the factor $Q_T$.

Proposition~\ref{prop:dla-regret} requires no assumption on the corrected costs. If, in addition, the corrected costs of an expert $k$ satisfy $\max_{a\in\A}|\widehat r^k_t(a)-r^*_t(a)|\leq\epsilon^k_t$, where $r^*_t(a)=\E[\ell(a,Y_t)/\ell_{\max}\mid\mathcal G_t]$ is the conditional risk of action $a$, then the expected loss of DLA exceeds that of the action minimizing $r^*_t$ by at most the expected second term of \eqref{eq:dla-regret} plus $2\ell_{\max}\E\sum_{t=1}^T\epsilon^k_t$ (Appendix~\ref{app:dla-conditional}). For the best response, $\epsilon^0_t\leq\frac12\|\widehat p_t-p_t^*\|_1$ by the argument of Proposition~\ref{prop:surrogate-validity}; the corrected experts can have smaller error when the predicted probabilities misstate the costs in a persistent way.

\subsection{Residual corrections as discounted least squares}
\label{app:dla-least-squares}

If a residual pool of DLA has received residuals $\xi_1,\ldots,\xi_n$, repeated application of \eqref{eq:dla-residual-update} gives $S(a)=\sum_{j=1}^n\rho_\tau^{n-j}\xi_j(a)$ and $N=\sum_{j=1}^n\rho_\tau^{n-j}$. The derivative of $\kappa u^2+\sum_{j=1}^n\rho_\tau^{n-j}\{\xi_j(a)-u\}^2$ with respect to $u$ is $2(\kappa+N)u-2S(a)$. Since $\kappa>0$, the objective is strictly convex, and its minimizer is $S(a)/(\kappa+N)$. Because $\ell(a,y)/\ell_{\max}$ and $r^0_s(a)$ lie in $[0,1]$, $|\xi_j(a)|\leq1$, so $|S(a)|\leq N$ and every correction lies in $(-1,1)$.

\subsection{Proof of Proposition~\ref{prop:dla-regret}}
\label{app:proof-dla-regret}

\begin{proof}
Fix a realized data stream and label arrival times. Let $\mathcal T_j$ be the set of rounds at which copy $j$ is used, and let $n_j=|\mathcal T_j|$, so that $\sum_{j=1}^{Q_T}n_j=T$. A copy is used again only after it has received the losses of its previous round. Hence, along $\mathcal T_j$ in increasing order, its weights are those of AdaHedge with immediate feedback on the loss vectors $(\lambda^k_t)_{k\in\mathcal E}$, $t\in\mathcal T_j$, and its loss at round $t\in\mathcal T_j$ is $\sum_kw^k_t\lambda^k_t=\bar\ell^{\rm DLA}_t/\ell_{\max}$. The experts change their actions between these rounds using the labels of rounds assigned to other copies. This does not affect the argument, since the bound of AdaHedge holds for every sequence of loss vectors in $[0,1]^M$. The label of the last round of a copy may arrive after round $T$; its loss vector is determined by $y_t$, and the bound below does not depend on when it arrives.

Applying Corollary~9 of \citet{derooij2014follow} to copy $j$ gives
\begin{equation*}
\sum_{t\in\mathcal T_j}\frac{\bar\ell^{\rm DLA}_t}{\ell_{\max}}
\leq
\min_{k\in\mathcal E}\sum_{t\in\mathcal T_j}\lambda_t^k
+\sqrt{n_j\ln M}+c_M,
\qquad c_M=\frac43\ln M+2,
\end{equation*}
because its loss coordinates lie in $[0,1]$.

For every fixed $k$, the minimum over each $\mathcal T_j$ is at most the loss of expert $k$ on $\mathcal T_j$, so $\sum_{j=1}^{Q_T}\min_{k\in\mathcal E}\sum_{t\in\mathcal T_j}\lambda^k_t\leq\min_{k\in\mathcal E}\sum_{t=1}^T\lambda^k_t$. By the Cauchy--Schwarz inequality, $\sum_{j=1}^{Q_T}\sqrt{n_j}\leq\sqrt{Q_T\sum_jn_j}=\sqrt{Q_TT}$. Summing the inequalities over $j$ and multiplying by $\ell_{\max}$ proves \eqref{eq:dla-regret}.

Each busy copy waits for exactly one label, and a copy is created only when all existing copies are busy. At that round, the number of existing copies is at most the number of unobserved labels of earlier rounds, so the number of copies after the creation is at most one plus that number.
\end{proof}

\subsection{Comparison of DLA with the best conditional action}
\label{app:dla-conditional}

Let $r^*_t(a)=\E[\ell(a,Y_t)/\ell_{\max}\mid\mathcal G_t]$, with $\mathcal G_t$ as in Appendix~\ref{subsec:dla-combination}, and let $a^*_t\in\argmin_{a\in\A}r^*_t(a)$. The chosen actions affect neither the data stream nor the predictive model, and neither the pools nor AdaHedge use them. For a fixed expert $k$, suppose $\max_{a\in\A}|\widehat r^k_t(a)-r^*_t(a)|\leq\epsilon^k_t$ for $\mathcal G_t$-measurable $\epsilon^k_t\geq0$. Because $a^k_t$ minimizes $\widehat r^k_t$,
$$
r^*_t(a^k_t)-r^*_t(a^*_t)
=\{r^*_t(a^k_t)-\widehat r^k_t(a^k_t)\}
+\{\widehat r^k_t(a^k_t)-\widehat r^k_t(a^*_t)\}
+\{\widehat r^k_t(a^*_t)-r^*_t(a^*_t)\}
\leq2\epsilon^k_t.
$$
Since $r^*_t(a)\in[0,1]$, the clipping in \eqref{eq:dla-corrected-policy} does not increase $|\widehat r^k_t(a)-r^*_t(a)|$. For the best response, with $p^*_t(y)=\mathbb P(Y_t=y\mid\mathcal G_t)$,
$$
|r^0_t(a)-r^*_t(a)|
=\left|\sum_{y\in\Y}(\widehat p_t(y)-p^*_t(y))
\left\{\frac{\ell(a,y)}{\ell_{\max}}-\frac12\right\}\right|
\leq\frac12\|\widehat p_t-p^*_t\|_1,
$$
as in the proof of Proposition~\ref{prop:surrogate-validity}. The actions $a^k_t$ and $a^*_t$ are $\mathcal G_t$-measurable, so $\E[\ell(a^k_t,Y_t)-\ell(a^*_t,Y_t)\mid\mathcal G_t]=\ell_{\max}\{r^*_t(a^k_t)-r^*_t(a^*_t)\}\leq2\ell_{\max}\epsilon^k_t$. Taking expectations in \eqref{eq:dla-regret}, bounding the minimum by the loss of expert $k$, and using $\E[\ell(a_t,Y_t)\mid\mathcal G_t,Y_t]=\bar\ell^{\rm DLA}_t$ gives
$$
\E\sum_{t=1}^T\left\{\ell(a_t,Y_t)-\ell(a^*_t,Y_t)\right\}
\leq
\ell_{\max}\,\E\left\{\sqrt{Q_TT\ln M}+Q_Tc_M\right\}
+2\ell_{\max}\min_{k\in\mathcal E}\E\sum_{t=1}^T\epsilon^k_t.
$$
The bound holds for every $k$, which gives the minimum over $\mathcal E$.

\section{Additional results}
\label{app:additional-results}

\subsection{Results for all cost ratios}
\label{app:all-cost-ratios}

Tables~\ref{tab:results-f005}--\ref{tab:results-f025} report all methods at $f=0.05$, $0.10$, and $0.25$, including the two runs that DCP combines, denoted DCP[$J$] and DCP[$V$], and Table~\ref{tab:paired-t} gives the paired $t$ statistics of DCP against each other method. At $f=0.05$, DCP reduces average decision loss relative to BR by $54.0\%$ on the AEMO dataset, $42.5\%$ on the CFPB dataset, and $2.8\%$ and $7.0\%$ on the Beijing and Flood datasets. At $f=0.25$, it reduces average decision loss by $20.0\%$ on the Flood dataset, $19.9\%$ on the AEMO dataset, and $2.8\%$ on the CFPB dataset, and increases it by $2.8\%$ on the Beijing dataset. 

Of the twelve settings, DLA has lower average decision loss than
DCP only on the Beijing dataset at $f=0.25$, where it is within $0.2\%$ of BR, and DCP has lower average decision loss than UP-$\widehat p$, ordinary ACI, and DtACI in all twelve. Relative to BR, DCP[$J$] is within $3.6\%$ on the Beijing dataset and reduces decision loss by $6.3\%$ to $8.1\%$ on the Flood dataset at every cost ratio. DCP[$V$] reduces decision loss by $21.5\%$ on the Flood dataset at $f=0.25$, where DCP[$J$] reduces it by $6.3\%$, and is $3.8\%$ to $24.0\%$ worse than BR on the Beijing dataset. In every setting, the decision loss of DCP exceeds that of the better of the two runs by at most $2.1\%$.

At $f=0.05$ on the binary datasets, the scalar robust action used with ordinary ACI and DtACI always protects. Indeed, write $k=(1-f)/f$. For $C=\emptyset,\{0\},\{1\},\{0,1\}$, the objective pairs for no protection and protection are respectively $(\alpha k,\alpha)$, $(\alpha k,1)$, $(k,\alpha)$, and $(k,1)$. Thus $\alpha k>1$, or equivalently $f<\alpha/(1+\alpha)$, makes protection the unique minimizer for every set. Their losses therefore equal the empirical non-event frequencies, $0.982$ on the AEMO dataset and $0.967$ on the CFPB dataset. 

\begin{table}
  \centering
  \footnotesize
  \setlength{\tabcolsep}{8pt}
  \caption{\footnotesize Results at $f=0.05$. DCP[$J$] and DCP[$V$] are Algorithm~\ref{alg:dcp} with the surrogate $J_t$ of \eqref{eq:J} and $V_t$ of \eqref{eq:VLW}, respectively; DCP combines the two by AdaHedge. DLA is Algorithm~\ref{alg:dla}; UP and UP-$\widehat p$ are described in Appendix~\ref{app:parameters}.}
  \label{tab:results-f005}
  \begin{tabular}{llccc}
    \toprule
    Dataset & Method & Coverage & Decision loss & Set size \\
    \midrule
      AEMO & BR & -- & $0.200$\se{0.023} & -- \\
       & DLA & -- & $0.107$\se{0.015} & -- \\
       & UP & -- & $0.216$\se{0.041} & -- \\
       & UP-$\widehat p$ & -- & $0.150$\se{0.030} & -- \\
       & Ordinary ACI & $0.905$ & $0.982$\se{0.006} & $0.94$ \\
       & DtACI & $0.909$ & $0.982$\se{0.006} & $0.94$ \\
       & DCP[$J$] & $0.900$ & $0.092$\se{0.015} & $0.92$ \\
       & DCP[$V$] & $0.900$ & $0.095$\se{0.016} & $0.92$ \\
       & DCP & $0.900$ & $0.092$\se{0.015} & $0.92$ \\
      \addlinespace
      CFPB & BR & -- & $0.535$\se{0.020} & -- \\
       & DLA & -- & $0.330$\se{0.015} & -- \\
       & UP & -- & $0.628$\se{0.042} & -- \\
       & UP-$\widehat p$ & -- & $0.335$\se{0.017} & -- \\
       & Ordinary ACI & $0.903$ & $0.967$\se{0.002} & $0.92$ \\
       & DtACI & $0.904$ & $0.967$\se{0.002} & $0.92$ \\
       & DCP[$J$] & $0.901$ & $0.308$\se{0.017} & $0.97$ \\
       & DCP[$V$] & $0.900$ & $0.307$\se{0.017} & $0.98$ \\
       & DCP & $0.901$ & $0.308$\se{0.017} & $0.97$ \\
      \addlinespace
      Beijing & BR & -- & $3.696$\se{0.119} & -- \\
       & DLA & -- & $3.700$\se{0.152} & -- \\
       & UP & -- & $6.099$\se{0.197} & -- \\
       & UP-$\widehat p$ & -- & $4.084$\se{0.190} & -- \\
       & Ordinary ACI & $0.901$ & $6.023$\se{0.196} & $1.80$ \\
       & DtACI & $0.902$ & $6.023$\se{0.196} & $1.82$ \\
       & DCP[$J$] & $0.900$ & $3.592$\se{0.149} & $2.14$ \\
       & DCP[$V$] & $0.900$ & $3.837$\se{0.172} & $2.12$ \\
       & DCP & $0.900$ & $3.593$\se{0.149} & $2.14$ \\
      \addlinespace
      Flood & BR & -- & $1.183$\se{0.528} & -- \\
       & DLA & -- & $1.180$\se{0.537} & -- \\
       & UP & -- & $5.605$\se{2.155} & -- \\
       & UP-$\widehat p$ & -- & $4.278$\se{2.314} & -- \\
       & Ordinary ACI & $0.912$ & $29.139$\se{0.606} & $0.94$ \\
       & DtACI & $0.905$ & $29.139$\se{0.606} & $1.00$ \\
       & DCP[$J$] & $0.900$ & $1.087$\se{0.498} & $0.94$ \\
       & DCP[$V$] & $0.899$ & $1.078$\se{0.492} & $0.95$ \\
       & DCP & $0.900$ & $1.100$\se{0.499} & $0.94$ \\
    \bottomrule
  \end{tabular}
\end{table}

\begin{table}
  \centering
  \footnotesize
  \setlength{\tabcolsep}{8pt}
  \caption{\footnotesize Results at $f=0.10$, in the format of Table~\ref{tab:results-f005}.}
  \label{tab:results-f010}
  \begin{tabular}{llccc}
    \toprule
    Dataset & Method & Coverage & Decision loss & Set size \\
    \midrule
      AEMO & BR & -- & $0.094$\se{0.014} & -- \\
       & DLA & -- & $0.061$\se{0.011} & -- \\
       & UP & -- & $0.124$\se{0.028} & -- \\
       & UP-$\widehat p$ & -- & $0.084$\se{0.018} & -- \\
       & Ordinary ACI & $0.905$ & $0.138$\se{0.012} & $0.94$ \\
       & DtACI & $0.909$ & $0.131$\se{0.010} & $0.94$ \\
       & DCP[$J$] & $0.900$ & $0.054$\se{0.009} & $0.92$ \\
       & DCP[$V$] & $0.900$ & $0.056$\se{0.011} & $0.92$ \\
       & DCP & $0.900$ & $0.055$\se{0.010} & $0.92$ \\
      \addlinespace
      CFPB & BR & -- & $0.244$\se{0.012} & -- \\
       & DLA & -- & $0.207$\se{0.011} & -- \\
       & UP & -- & $0.298$\se{0.020} & -- \\
       & UP-$\widehat p$ & -- & $0.215$\se{0.014} & -- \\
       & Ordinary ACI & $0.903$ & $0.212$\se{0.015} & $0.92$ \\
       & DtACI & $0.904$ & $0.213$\se{0.015} & $0.92$ \\
       & DCP[$J$] & $0.900$ & $0.201$\se{0.011} & $0.97$ \\
       & DCP[$V$] & $0.900$ & $0.203$\se{0.012} & $0.93$ \\
       & DCP & $0.900$ & $0.202$\se{0.011} & $0.97$ \\
      \addlinespace
      Beijing & BR & -- & $2.852$\se{0.096} & -- \\
       & DLA & -- & $2.823$\se{0.124} & -- \\
       & UP & -- & $5.850$\se{0.242} & -- \\
       & UP-$\widehat p$ & -- & $3.086$\se{0.136} & -- \\
       & Ordinary ACI & $0.901$ & $3.248$\se{0.154} & $1.80$ \\
       & DtACI & $0.902$ & $3.264$\se{0.150} & $1.82$ \\
       & DCP[$J$] & $0.900$ & $2.748$\se{0.111} & $2.03$ \\
       & DCP[$V$] & $0.900$ & $3.049$\se{0.139} & $1.99$ \\
       & DCP & $0.900$ & $2.749$\se{0.110} & $2.03$ \\
      \addlinespace
      Flood & BR & -- & $0.834$\se{0.371} & -- \\
       & DLA & -- & $0.823$\se{0.383} & -- \\
       & UP & -- & $3.061$\se{1.389} & -- \\
       & UP-$\widehat p$ & -- & $2.300$\se{1.222} & -- \\
       & Ordinary ACI & $0.912$ & $2.849$\se{0.624} & $0.94$ \\
       & DtACI & $0.905$ & $3.665$\se{0.535} & $1.00$ \\
       & DCP[$J$] & $0.899$ & $0.781$\se{0.361} & $0.95$ \\
       & DCP[$V$] & $0.899$ & $0.760$\se{0.355} & $0.94$ \\
       & DCP & $0.899$ & $0.769$\se{0.359} & $0.95$ \\
    \bottomrule
  \end{tabular}
\end{table}

\begin{table}
  \centering
  \footnotesize
  \setlength{\tabcolsep}{8pt}
  \caption{\footnotesize Results at $f=0.25$, in the format of Table~\ref{tab:results-f005}.}
  \label{tab:results-f025}
  \begin{tabular}{llccc}
    \toprule
    Dataset & Method & Coverage & Decision loss & Set size \\
    \midrule
      AEMO & BR & -- & $0.032$\se{0.008} & -- \\
       & DLA & -- & $0.029$\se{0.008} & -- \\
       & UP & -- & $0.049$\se{0.013} & -- \\
       & UP-$\widehat p$ & -- & $0.036$\se{0.009} & -- \\
       & Ordinary ACI & $0.905$ & $0.120$\se{0.009} & $0.94$ \\
       & DtACI & $0.909$ & $0.113$\se{0.007} & $0.94$ \\
       & DCP[$J$] & $0.900$ & $0.028$\se{0.008} & $0.92$ \\
       & DCP[$V$] & $0.900$ & $0.026$\se{0.007} & $0.91$ \\
       & DCP & $0.900$ & $0.026$\se{0.007} & $0.92$ \\
      \addlinespace
      CFPB & BR & -- & $0.102$\se{0.006} & -- \\
       & DLA & -- & $0.099$\se{0.006} & -- \\
       & UP & -- & $0.099$\se{0.007} & -- \\
       & UP-$\widehat p$ & -- & $0.099$\se{0.007} & -- \\
       & Ordinary ACI & $0.903$ & $0.113$\se{0.004} & $0.92$ \\
       & DtACI & $0.904$ & $0.113$\se{0.004} & $0.92$ \\
       & DCP[$J$] & $0.900$ & $0.101$\se{0.006} & $0.96$ \\
       & DCP[$V$] & $0.900$ & $0.099$\se{0.006} & $0.92$ \\
       & DCP & $0.900$ & $0.099$\se{0.006} & $0.96$ \\
      \addlinespace
      Beijing & BR & -- & $1.783$\se{0.081} & -- \\
       & DLA & -- & $1.786$\se{0.087} & -- \\
       & UP & -- & $4.442$\se{0.291} & -- \\
       & UP-$\widehat p$ & -- & $1.935$\se{0.088} & -- \\
       & Ordinary ACI & $0.901$ & $1.934$\se{0.099} & $1.80$ \\
       & DtACI & $0.902$ & $1.992$\se{0.089} & $1.82$ \\
       & DCP[$J$] & $0.900$ & $1.832$\se{0.091} & $1.96$ \\
       & DCP[$V$] & $0.900$ & $2.211$\se{0.118} & $1.94$ \\
       & DCP & $0.900$ & $1.834$\se{0.091} & $1.96$ \\
      \addlinespace
      Flood & BR & -- & $0.435$\se{0.209} & -- \\
       & DLA & -- & $0.405$\se{0.193} & -- \\
       & UP & -- & $1.326$\se{0.709} & -- \\
       & UP-$\widehat p$ & -- & $0.958$\se{0.525} & -- \\
       & Ordinary ACI & $0.912$ & $2.580$\se{0.560} & $0.94$ \\
       & DtACI & $0.905$ & $3.373$\se{0.463} & $1.00$ \\
       & DCP[$J$] & $0.900$ & $0.407$\se{0.197} & $0.95$ \\
       & DCP[$V$] & $0.902$ & $0.341$\se{0.162} & $0.92$ \\
       & DCP & $0.900$ & $0.348$\se{0.166} & $0.95$ \\
    \bottomrule
  \end{tabular}
\end{table}

\begin{table}
  \centering
  \footnotesize
  \setlength{\tabcolsep}{8pt}
  \caption{\footnotesize Paired $t$ statistics of the decision loss of DCP minus that of each other method, computed with the stationary bootstrap of Appendix~\ref{app:standard-errors}. Negative values favor DCP.}
  \label{tab:paired-t}
  \begin{tabular}{llcccccc}
    \toprule
    Dataset & $f$ & BR & DLA & UP & UP-$\widehat p$ & Ordinary ACI & DtACI \\
    \midrule
      AEMO & $0.05$ & $-6.6$ & $-3.8$ & $-4.0$ & $-3.0$ & $-45.9$ & $-45.9$ \\
       & $0.10$ & $-4.9$ & $-2.0$ & $-3.4$ & $-3.0$ & $-11.3$ & $-14.5$ \\
       & $0.25$ & $-2.4$ & $-1.9$ & $-3.1$ & $-3.5$ & $-14.5$ & $-18.9$ \\
      CFPB & $0.05$ & $-10.2$ & $-5.5$ & $-11.5$ & $-2.9$ & $-35.4$ & $-35.4$ \\
       & $0.10$ & $-3.2$ & $-3.2$ & $-10.3$ & $-4.1$ & $-2.4$ & $-2.7$ \\
       & $0.25$ & $-4.0$ & $-2.0$ & $-0.6$ & $-0.3$ & $-5.9$ & $-6.1$ \\
      Beijing & $0.05$ & $-1.2$ & $-1.3$ & $-10.1$ & $-2.9$ & $-10.0$ & $-10.0$ \\
       & $0.10$ & $-2.6$ & $-1.9$ & $-11.2$ & $-4.0$ & $-5.9$ & $-6.2$ \\
       & $0.25$ & $1.6$ & $1.6$ & $-9.1$ & $-2.3$ & $-2.7$ & $-3.5$ \\
      Flood & $0.05$ & $-1.3$ & $-1.7$ & $-2.7$ & $-1.7$ & $-26.2$ & $-26.2$ \\
       & $0.10$ & $-2.3$ & $-1.4$ & $-2.2$ & $-1.7$ & $-3.8$ & $-6.1$ \\
       & $0.25$ & $-1.6$ & $-1.6$ & $-1.7$ & $-1.7$ & $-4.2$ & $-6.5$ \\
    \bottomrule
  \end{tabular}
\end{table}

\subsection{Realized losses of the randomized actions}
\label{app:realized}

For DCP and DLA, Table~\ref{tab:realized} gives the mean and the $2.5\%$ and $97.5\%$ quantiles of the average realized loss over $1{,}000$ independent draws of the actions from their AdaHedge weights. In every setting, the mean differs from the expected loss under these weights, reported in the other tables, by less than $0.001$.

\begin{table}
  \centering
  \footnotesize
  \setlength{\tabcolsep}{4pt}
  \caption{\footnotesize Average decision loss of DCP and DLA: the expected loss under the AdaHedge weights, and the mean and the $2.5\%$ and $97.5\%$ quantiles of the average realized loss over $1{,}000$ independent draws of the actions.}
  \label{tab:realized}
  \begin{tabular}{llcccccc}
    \toprule
    & & \multicolumn{3}{c}{DCP} & \multicolumn{3}{c}{DLA} \\
    \cmidrule(lr){3-5}\cmidrule(lr){6-8}
    Dataset & $f$ & Expected & Realized & Range & Expected & Realized & Range \\
    \midrule
      AEMO & $0.05$ & $0.0920$ & $0.0920$ & $[0.0919,0.0922]$ & $0.1074$ & $0.1075$ & $[0.1045,0.1105]$ \\
       & $0.10$ & $0.0546$ & $0.0546$ & $[0.0537,0.0558]$ & $0.0610$ & $0.0610$ & $[0.0594,0.0627]$ \\
       & $0.25$ & $0.0258$ & $0.0258$ & $[0.0257,0.0259]$ & $0.0290$ & $0.0290$ & $[0.0282,0.0298]$ \\
      CFPB & $0.05$ & $0.3076$ & $0.3076$ & $[0.3071,0.3080]$ & $0.3301$ & $0.3302$ & $[0.3284,0.3320]$ \\
       & $0.10$ & $0.2018$ & $0.2018$ & $[0.2014,0.2022]$ & $0.2075$ & $0.2075$ & $[0.2063,0.2087]$ \\
       & $0.25$ & $0.0986$ & $0.0986$ & $[0.0986,0.0987]$ & $0.0992$ & $0.0992$ & $[0.0987,0.0996]$ \\
      Beijing & $0.05$ & $3.5926$ & $3.5926$ & $[3.5925,3.5927]$ & $3.7004$ & $3.7005$ & $[3.6768,3.7244]$ \\
       & $0.10$ & $2.7492$ & $2.7492$ & $[2.7487,2.7498]$ & $2.8233$ & $2.8230$ & $[2.8099,2.8355]$ \\
       & $0.25$ & $1.8337$ & $1.8337$ & $[1.8333,1.8340]$ & $1.7857$ & $1.7859$ & $[1.7803,1.7917]$ \\
      Flood & $0.05$ & $1.1002$ & $1.1001$ & $[1.0678,1.1185]$ & $1.1797$ & $1.1794$ & $[1.1632,1.1945]$ \\
       & $0.10$ & $0.7691$ & $0.7691$ & $[0.7580,0.7794]$ & $0.8228$ & $0.8219$ & $[0.7658,0.8437]$ \\
       & $0.25$ & $0.3475$ & $0.3477$ & $[0.3410,0.3585]$ & $0.4049$ & $0.4048$ & $[0.3917,0.4170]$ \\
    \bottomrule
  \end{tabular}
\end{table}

\subsection{Ablations of the action rule and the level selection}
\label{app:uncalibrated}

We compare the corrected labelwise action rule \eqref{eq:corrected-action} with the scalar robust rule \eqref{eq:robust action}.

The scalar action variant sets $d_t\equiv0$ and replaces \eqref{eq:corrected-action} by $a_t^{\rm sc}(\beta)\in\argmin_{a\in\A}L(C_{t,\beta},a;\beta)$. Thus $\beta$ indexes the candidate set and also serves as its scalar out-of-set charge. The scalar $J$ variant evaluates this action under $\widehat p_t$. The scalar $V$ variant uses the normalized minimum $V_t^{\rm sc}(\beta)=\min_a L(C_{t,\beta},a;\beta)/\ell_{\max}$. The other update rules are unchanged.

\begin{table}
  \centering
  \footnotesize
  \setlength{\tabcolsep}{8pt}
  \caption{\footnotesize Scalar robust action variants at $f=0.10$, in the format of Table~\ref{tab:main-results}. The ordinary ACI, DCP[$J$], and DCP[$V$] rows are repeated from Table~\ref{tab:results-f010} for comparison.}
  \label{tab:uncalibrated}
  \begin{tabular}{llccc}
    \toprule
    Dataset & Method & Coverage & Decision loss & Set size \\
    \midrule
      AEMO & Ordinary ACI & $0.905$ & $0.138$\se{0.012} & $0.94$ \\
       & $J$, scalar & $0.900$ & $0.152$\se{0.011} & $0.92$ \\
       & DCP[$J$] & $0.900$ & $0.054$\se{0.009} & $0.92$ \\
       & $V$, scalar & $0.900$ & $0.140$\se{0.012} & $0.91$ \\
       & DCP[$V$] & $0.900$ & $0.056$\se{0.011} & $0.92$ \\
      \addlinespace
      CFPB & Ordinary ACI & $0.903$ & $0.212$\se{0.015} & $0.92$ \\
       & $J$, scalar & $0.900$ & $0.244$\se{0.012} & $0.96$ \\
       & DCP[$J$] & $0.900$ & $0.201$\se{0.011} & $0.97$ \\
       & $V$, scalar & $0.900$ & $0.211$\se{0.015} & $0.92$ \\
       & DCP[$V$] & $0.900$ & $0.203$\se{0.012} & $0.93$ \\
      \addlinespace
      Beijing & Ordinary ACI & $0.901$ & $3.248$\se{0.154} & $1.80$ \\
       & $J$, scalar & $0.900$ & $2.952$\se{0.093} & $2.03$ \\
       & DCP[$J$] & $0.900$ & $2.748$\se{0.111} & $2.03$ \\
       & $V$, scalar & $0.900$ & $4.066$\se{0.177} & $2.44$ \\
       & DCP[$V$] & $0.900$ & $3.049$\se{0.139} & $1.99$ \\
      \addlinespace
      Flood & Ordinary ACI & $0.912$ & $2.849$\se{0.624} & $0.94$ \\
       & $J$, scalar & $0.904$ & $3.157$\se{0.326} & $0.94$ \\
       & DCP[$J$] & $0.899$ & $0.781$\se{0.361} & $0.95$ \\
       & $V$, scalar & $0.900$ & $3.422$\se{0.322} & $0.96$ \\
       & DCP[$V$] & $0.899$ & $0.760$\se{0.355} & $0.94$ \\
    \bottomrule
  \end{tabular}
\end{table}

Table~\ref{tab:uncalibrated} reports the comparison at $f=0.10$. The corrected labelwise action rule has lower average loss than the scalar robust rule on every dataset under both surrogates: the reductions range from $6.9\%$ to $75.2\%$ under $J_t$ and from $3.9\%$ to $77.8\%$ under $V_t$. The scalar $J$ variant is worse than ordinary ACI on the AEMO, CFPB, and Flood datasets and better on the Beijing dataset; the scalar $V$ variant has approximately the same loss as ordinary ACI on the AEMO and CFPB datasets and higher loss on the Beijing and Flood datasets. Under $V_t$ on the Beijing dataset, the labelwise rule also reduces the average set size from $2.44$ to $1.99$. The comparisons at $f=0.05$ and $f=0.25$ likewise favor the labelwise rule in all sixteen settings.

Two further comparisons separate the action rule from the selection of the level. In the first, we keep the sets of ordinary ACI and replace its scalar robust rule by the DCP action rule \eqref{eq:corrected-action}, with the out-of-set masses of the ACI set formed and corrected as in Section~\ref{subsec:surrogate}. The sets, and hence the coverage, are those of ordinary ACI. The average decision loss is lower than that of ordinary ACI in all twelve settings, at $f=0.10$ by $49.9\%$ on the AEMO dataset, $5.2\%$ on the CFPB dataset, $13.0\%$ on the Beijing dataset, and $72.8\%$ on the Flood dataset. The paired $t$ statistics against ordinary ACI are below $-1.96$ in eleven settings, and $-1.9$ on the Beijing dataset at $f=0.25$.

In the second, we run DCP with the single level $\B=\{\alpha\}$ in place of the grid, so that no level is selected and the runs with $J_t$ and $V_t$ coincide. On the Beijing dataset, the average decision loss is $3.849$, $3.053$, and $2.213$ at $f=0.05$, $0.10$, and $0.25$, against $3.593$, $2.749$, and $1.834$ for DCP, and the paired $t$ statistics of DCP against this variant are $-3.3$, $-4.4$, and $-5.3$. On the other three datasets, the two losses differ by at most $3\%$, with paired $t$ statistics between $-1.0$ and $1.5$. These comparisons indicate that the reduction relative to ordinary ACI comes mainly from the DCP action rule with corrected out-of-set masses, and partly from the selection of the level.

\subsection{Other regions of the electricity market}
\label{app:regions}

Table~\ref{tab:aemo-regions} repeats the AEMO experiment for the other four regions of the market, with the same features; the neighbouring region is Queensland for New South Wales, New South Wales for Queensland and Victoria, and Victoria for Tasmania. In every region and at every cost ratio, DCP has lower average decision loss than BR, by $6.9\%$ (Tasmania, $f=0.25$) to $76.0\%$ (Queensland, $f=0.05$), and than ordinary ACI, by $42.6\%$ (New South Wales, $f=0.10$) to $95.1\%$ (Tasmania, $f=0.05$).

\begin{table}
  \centering
  \footnotesize
  \setlength{\tabcolsep}{4pt}
  \caption{\footnotesize Average decision loss of ordinary ACI, BR, and DCP in the five regions of the electricity market.}
  \label{tab:aemo-regions}
  \begin{tabular}{lccccccccc}
    \toprule
    & \multicolumn{3}{c}{$f=0.05$} & \multicolumn{3}{c}{$f=0.10$} & \multicolumn{3}{c}{$f=0.25$} \\
    \cmidrule(lr){2-4}\cmidrule(lr){5-7}\cmidrule(lr){8-10}
    Region & ACI & BR & DCP & ACI & BR & DCP & ACI & BR & DCP \\
    \midrule
      SA & $0.982$ & $0.200$ & $0.092$ & $0.138$ & $0.094$ & $0.055$ & $0.120$ & $0.032$ & $0.026$ \\
      NSW & $0.986$ & $0.316$ & $0.154$ & $0.148$ & $0.167$ & $0.085$ & $0.125$ & $0.058$ & $0.039$ \\
      QLD & $0.993$ & $0.263$ & $0.063$ & $0.112$ & $0.125$ & $0.044$ & $0.107$ & $0.036$ & $0.020$ \\
      VIC & $0.989$ & $0.086$ & $0.055$ & $0.130$ & $0.050$ & $0.034$ & $0.122$ & $0.019$ & $0.016$ \\
      TAS & $0.995$ & $0.123$ & $0.048$ & $0.172$ & $0.045$ & $0.027$ & $0.165$ & $0.011$ & $0.011$ \\
    \bottomrule
  \end{tabular}
\end{table}

\subsection{Label delay on the CFPB dataset}
\label{app:cfpb-delay}

In the experiments above, the label of a CFPB complaint is observed $15$ days after the complaint was sent to the company, or $60$ days for an untimely response (Appendix~\ref{app:datasets}). As a check, we repeat the CFPB experiment with every label observed $60$ days after the complaint was sent to the company, and with the step-size caps of Appendix~\ref{app:delays} computed with $D=60$. The average decision loss of DCP is then $0.317$, $0.201$, and $0.099$ at $f=0.05$, $0.10$, and $0.25$, against $0.308$, $0.202$, and $0.099$ in Tables~\ref{tab:results-f005}--\ref{tab:results-f025}, with coverage $0.900$. It remains lower than that of each of the six compared methods at every cost ratio. The paired $t$ statistics of DCP against these methods are below $-1.96$ at $f=0.05$ and $0.10$. At $f=0.25$, they are $-1.0$ against DLA and $-1.2$ against UP-$\widehat p$, compared with $-2.0$ and $-0.3$ in Table~\ref{tab:paired-t}, and below $-1.96$ against the other four methods.

\subsection{Miscalibration on the AEMO and CFPB datasets}
\label{app:miscalibration}

The binary cost--loss problems make the effect of probability errors on actions explicit. At $f=0.10$, the protective action has lower expected cost when the event probability exceeds $0.10$, and BR acts whenever the predicted probability crosses this threshold. We examine temporal changes in event frequency and errors near the threshold to understand the losses reported in Section~\ref{sec:results-headline}.

On the AEMO dataset, spikes occur in $14.2\%$ of half-hour intervals in 2022, $3.4\%$ in 2023, $2.6\%$ in 2024, and $1.6\%$ in 2025. The training set ends on 14 March 2025, and $70\%$ of its spikes occurred in 2022. Predicted probabilities overestimate the event frequency in every test-set calendar quarter: the average prediction and observed frequency are respectively $1.4\%$ and $0.2\%$ from 14 to 31 March 2025, $7.8\%$ and $2.4\%$ from April to June, $5.0\%$ and $2.9\%$ from July to September, and $0.4\%$ and $0.3\%$ from October to December.

CFPB shows a related pattern. The average predicted probability of monetary relief remains between $6.1\%$ and $8.4\%$ in each test year, while the fraction of complaints closed with monetary relief declines from $4.5\%$ in 2019 to $1.9\%$ in 2024.

Table~\ref{tab:miscalibration-errors} reports protective actions, false alarms, and missed events. DLA and DCP act about half as often as BR: in $4.3\%$ and $3.8\%$ of cases against $8.8\%$ on the AEMO dataset, and in $11.3\%$ and $9.7\%$ against $19.9\%$ on the CFPB dataset. Relative to BR, this reduces false alarms but increases missed events; under the specified loss, the reduction in false alarms more than offsets the additional misses. Comparing the adaptive methods with each other, DCP acts less often than DLA, with $48$ versus $50$ missed events on the AEMO dataset and $2{,}805$ versus $2{,}594$ on the CFPB dataset. Its average decision loss is lower on both datasets.

Table~\ref{tab:miscalibration-bands} locates most of the difference between DLA and DCP near the decision threshold. On the AEMO dataset, observations with predicted probability in $[0.10,0.20)$ account for $60\%$ of the loss difference; on the CFPB dataset, those in $[0.10,0.15)$ account for $100\%$. In these ranges the event frequency is below $10\%$, yet DLA takes the protective action more often than DCP. The loss difference is also largest during periods with pronounced overestimation: DLA exceeds DCP by $0.006$ and $0.014$ per case in the second and third quarters of 2025 on the AEMO dataset, and by $0.012$ to $0.027$ per case from the second quarter of 2022 through the first quarter of 2023 on the CFPB dataset. In the latter period, monetary relief occurs in $2.2\%$--$3.0\%$ of cases while predicted probabilities average $8.2\%$--$8.8\%$. These observations support the interpretation that DCP better adjusts the relevant decisions to the observed miscalibration in these datasets.

\begin{table}
  \centering
  \footnotesize
  \setlength{\tabcolsep}{6pt}
  \caption{\footnotesize Protective actions on the AEMO and CFPB datasets at $f=0.10$: share of cases with the protective action, false alarms, missed events, and average decision loss. DLA and DCP are counted in expectation under their AdaHedge weights.}
  \label{tab:miscalibration-errors}
  \begin{tabular}{llcccc}
    \toprule
    Dataset & Method & Action & False alarms & Missed events & Decision loss \\
    \midrule
      AEMO & BR & $8.8\%$ & $1{,}017$ & $34$ & $0.094$ \\
       & DLA & $4.3\%$ & $403$ & $50$ & $0.061$ \\
       & DCP & $3.8\%$ & $333$ & $48$ & $0.055$ \\
      CFPB & BR & $19.9\%$ & $35{,}288$ & $1{,}600$ & $0.244$ \\
       & DLA & $11.3\%$ & $18{,}850$ & $2{,}594$ & $0.207$ \\
       & DCP & $9.7\%$ & $15{,}796$ & $2{,}805$ & $0.202$ \\
    \bottomrule
  \end{tabular}
\end{table}

\begin{table}
  \centering
  \footnotesize
  \setlength{\tabcolsep}{4pt}
  \caption{\footnotesize Share of cases with the protective action on the AEMO and CFPB datasets at $f=0.10$, by range of predicted probability, with the number of cases and the observed frequency of the event in each range. The last column gives the share of the total difference between the decision losses of DLA and DCP that arises in each range.}
  \label{tab:miscalibration-bands}
  \begin{tabular}{llcccccc}
    \toprule
    Dataset & Predicted probability & Cases & Observed & BR & DLA & DCP & Difference \\
    \midrule
      AEMO & $[0.00,0.10)$ & $12{,}794$ & $0.3\%$ & $0.0\%$ & $0.0\%$ & $0.1\%$ & $19\%$ \\
       & $[0.10,0.15)$ & $426$ & $3.5\%$ & $100.0\%$ & $10.5\%$ & $5.5\%$ & $15\%$ \\
       & $[0.15,0.20)$ & $257$ & $5.4\%$ & $100.0\%$ & $26.9\%$ & $17.1\%$ & $45\%$ \\
       & $[0.20,0.30)$ & $222$ & $12.2\%$ & $100.0\%$ & $72.2\%$ & $56.4\%$ & $17\%$ \\
       & $[0.30,1]$ & $325$ & $48.3\%$ & $100.0\%$ & $100.0\%$ & $99.1\%$ & $3\%$ \\
      CFPB & $[0.00,0.10)$ & $162{,}887$ & $1.0\%$ & $0.0\%$ & $0.0\%$ & $0.0\%$ & $1\%$ \\
       & $[0.10,0.15)$ & $19{,}088$ & $5.2\%$ & $99.7\%$ & $18.9\%$ & $3.2\%$ & $100\%$ \\
       & $[0.15,0.20)$ & $7{,}130$ & $13.9\%$ & $100.0\%$ & $72.4\%$ & $68.0\%$ & $-4\%$ \\
       & $[0.20,0.30)$ & $7{,}343$ & $19.5\%$ & $100.0\%$ & $99.3\%$ & $100.0\%$ & $3\%$ \\
       & $[0.30,1]$ & $6{,}909$ & $24.8\%$ & $100.0\%$ & $100.0\%$ & $100.0\%$ & $0\%$ \\
    \bottomrule
  \end{tabular}
\end{table}

\subsection{Trajectories on the other datasets}
\label{app:trajectories}

\begin{figure}
\centering
\safeincludegraphics[width=0.8\textwidth]{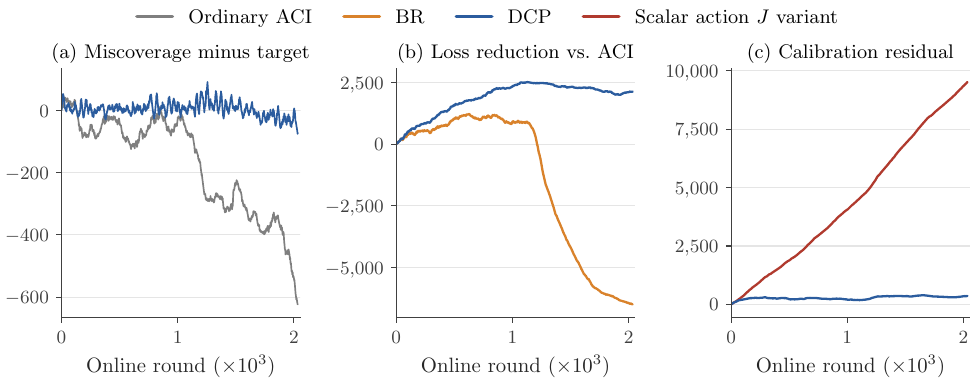}
\caption{\footnotesize The panels of Figure~\ref{fig:trajectory-aemo} for CFPB at $f=0.10$, a single run.}
\label{fig:traj-cfpb}
\end{figure}

\begin{figure}
\centering
\safeincludegraphics[width=0.8\textwidth]{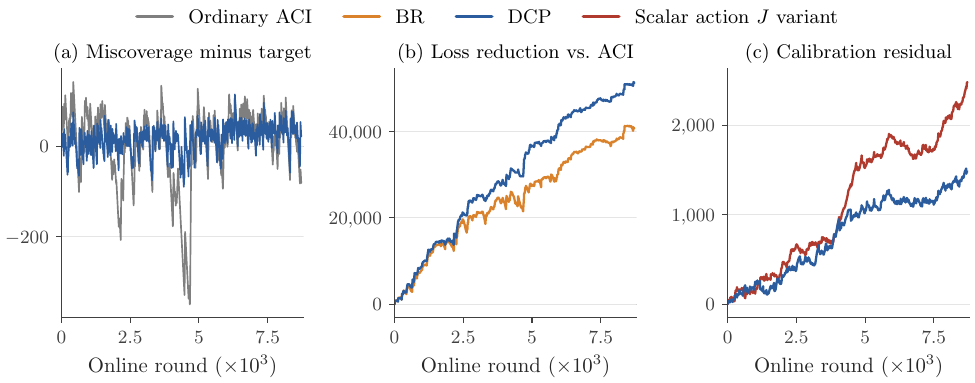}
\caption{\footnotesize The panels of Figure~\ref{fig:trajectory-aemo} for Beijing at $f=0.10$, a single run.}
\label{fig:traj-beijing}
\end{figure}

\begin{figure}
\centering
\safeincludegraphics[width=0.8\textwidth]{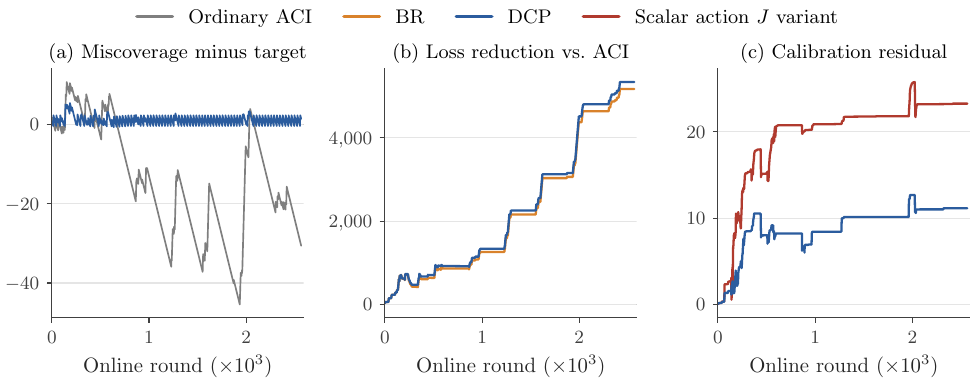}
\caption{\footnotesize The panels of Figure~\ref{fig:trajectory-aemo} for Flood at $f=0.10$, a single run.}
\label{fig:traj-flood}
\end{figure}

Figures~\ref{fig:traj-cfpb}--\ref{fig:traj-flood} repeat Figure~\ref{fig:trajectory-aemo} for the other three datasets at $f=0.10$, one run each; on the CFPB and Beijing datasets, the horizontal axis counts days and hours.

\end{document}